\documentclass[11pt]{article}

\usepackage[final]{acl}

\usepackage{times}
\usepackage{latexsym}

\usepackage[T1]{fontenc}

\usepackage[utf8]{inputenc}

\usepackage{microtype}

\usepackage{inconsolata}

\usepackage{graphicx}
\usepackage{enumitem}
\usepackage{xspace}
\usepackage{adjustbox}
\usepackage{booktabs}
\usepackage{multirow}
\usepackage{makecell}
\usepackage[table,xcdraw]{xcolor}
\usepackage{colortbl}
\usepackage{algpseudocode}
\usepackage{algorithm}
\usepackage{caption}

\usepackage{amsfonts}
\usepackage{graphicx}
\usepackage{pifont}
\usepackage{xcolor}
\usepackage{fontawesome}
\usepackage{makecell}
\usepackage{caption}
\usepackage{subcaption}
\usepackage{amsmath}
\usepackage{amsfonts}
\usepackage{amsbsy}
\usepackage{amsthm}

\usepackage{amsmath}
\usepackage{amsfonts}
\usepackage{amsbsy}
\usepackage{amsthm}
\usepackage{MnSymbol}

\usepackage[most]{tcolorbox}

\title{Zero-Shot Tabular Learning via Hypergraph Reasoning with LLMs}

\author{
\begin{tabular}{
  c@{\hspace{2.0em}}
  c@{\hspace{2.0em}}
  c
}
\textbf{Geon Lee} &
\textbf{Kyungho Kim} &
\textbf{Dongwon Choi} \\[3pt]
\textbf{Juyeon Kim} &
\textbf{Shinhwan Kang} &
\textbf{Kijung Shin}
\end{tabular}
\\[5pt]
KAIST
\\[2pt]
\texttt{\{geonlee0325, kkyungho, cookie000215, juyeonkim,}\\[-1pt]
\texttt{shinhwan.kang, kijungs\}@kaist.ac.kr}
}

\begin{document}

\newcommand\red[1]{\textcolor{red}{#1}}
\newcommand\blue[1]{\textcolor{blue}{#1}}

\newcommand{\cond}{\textsc{No-Prop.}\xspace}

\newcommand{\pseudo}{\textsc{TabHGR}\textsuperscript{-}\xspace}
\newcommand{\naive}{\textsc{TabHGR}\xspace}
\newcommand{\method}{\textsc{TabHGR}\textsuperscript{+}\xspace}

\newcommand{\smallsection}[1]{{\vspace{0.05in} \noindent {\bf{\underline{\smash{#1}}}}}}

\definecolor{PseudoFalse}{RGB}{238, 172, 172}  
\definecolor{PseudoTrue}{RGB}{160, 191, 213}  

\maketitle

\begin{abstract}

Tabular data are prevalent across diverse domains such as healthcare and finance.
Zero-shot tabular learning aims to make predictions without labeled data.
However, existing LLM-based approaches rely on row-wise prompting, resulting in high inference cost and limited use of inter-row dependencies.
In this work, we propose \method, an efficient relation-aware zero-shot framework for tabular data that leverages LLM-generated rules to label and group rows.
Specifically, based on column and task descriptions, we use LLMs to generate rules that assign weak labels to all rows.
To refine these noisy weak labels, we construct a hypergraph by grouping semantically related rows using LLM-generated rules and perform hypergraph-based label propagation.
Lastly, we further enhance the LLM-generated hypergraph for propagation, which is potentially suboptimal, by assigning label-adaptive weights to hyperedges.
By requiring only a constant number of LLM calls independent of dataset size, \method substantially reduces inference cost while achieving higher classification accuracy than zero-shot and strong few-shot baselines.

\end{abstract}

\section{Introduction}
\label{sec:intro}

Tabular data are ubiquitous across real-world domains such as healthcare, finance, and material science. 
Despite their prevalence, effective learning from tabular data often relies on labeled supervision, which is expensive, unavailable, or impractical to obtain. 
These challenges motivate \textit{zero-shot tabular learning}, which aims to perform predictions on a new task without labeled examples or task-specific fine-tuning.


Large language models (LLMs), pretrained on broad world knowledge, offer strong potential for zero-shot reasoning.
Recent work applies LLMs to tabular data by prompting the model with individual rows (tuples)~\cite{slack2023tablet,hegselmann2023tabllm}.
However, such \textit{row-wise} reasoning relies on per-row prompts and incurs substantial inference cost that grows with the dataset size.
In addition, it overlooks \textit{inter-row} dependencies, which are prevalent in real-world tabular data.
In practice, rows are often semantically related, such as patients with similar clinical profiles or financial records sharing risk patterns, which are not fully captured by row-wise predictions.

In this work, we explore a new way of leveraging LLMs in zero-shot tabular learning to address these limitations.
Rather than prompting LLMs on a per-row basis, we use them to generate attribute (column)-based rules not just for weak labeling but also for grouping related rows, which together enable efficient, relation-aware reasoning.



Specifically, we first use an LLM to produce a small set of attribute-based rules that characterize each class.
These rules are applied to all rows to produce initial \textit{weak labels}.
This requires only a constant number of LLM prompts based on attribute and class descriptions, independent of dataset size.
However, the resulting weak labels are inevitably noisy, as they rely on coarse rules.

To refine these noisy labels, we then exploit inter-row relations that are not explicitly given but are inferred by LLMs.
Specifically, we use LLMs to produce attribute-based rules that group related rows based on their semantic understanding of attributes.
These groups, which correspond to \textit{hyperedges}, together form a \textit{hypergraph} that models higher-order relations among rows.
Weak labels are then propagated over this hypergraph, refining them by aggregating information while accounting for these higher-order relations.
This propagation requires only parameter-free forward computations on the hypergraph, without gradient-based optimization.


Lastly, we further enhance the LLM-generated hypergraph for propagation, which may be suboptimal, by assigning hyperedge weights. 
These weights are label-adaptive, motivated by our observation that the effectiveness of a hyperedge varies across weak labels.
Specifically, we observe that some hyperedges facilitate the propagation of certain weak labels while degrading others.
By selectively emphasizing hyperedges that are beneficial for each weak label, we achieve more effective weak label aggregation, leading to improved zero-shot prediction accuracy.


These ideas form \method, our proposed method for zero-shot tabular learning. Experiments on 9 benchmarks show that \method outperforms recent zero-shot methods and strong few-shot baselines, while requiring substantially lower inference cost than row-wise LLM prediction.

Our contributions are summarized as follows:
\begin{itemize}[leftmargin=*, noitemsep, topsep=0pt]
    \item \textbf{Novel usage of LLMs.}
    For zero-shot tabular learning, we use LLMs to produce attribute-based rules not just for labeling but also for grouping,
    with constant inference cost independent of dataset size.
    
    \item \textbf{Method.}
    We propose \method, a zero-shot tabular learning framework that integrates LLM-generated weak labels and higher-order inter-row relations through hypergraph-based propagation with adaptive weights.

    \item \textbf{Experiments.} 
    We conduct extensive experiments on 9 benchmark datasets and demonstrate that \method outperforms zero-shot LLM-based methods and strong few-shot baselines.
\end{itemize}
Our code and datasets are available at \url{https://github.com/geon0325/TabHGR}.
All code is hosted openly for academic use.

\section{Related Work}
\label{sec:related}
We review prior work relevant to this study.

\smallsection{LLMs for Tabular Learning.}
LLMs enable rich semantic reasoning over attributes, opening new opportunities for tabular learning.
Most existing approaches apply LLMs by serializing each row into natural language prompts for direct prediction or fine-tuning~\cite{hegselmann2023tabllm,dinh2022lift,nam2023semi}, or by deriving interpretable class-level rules and rationales applicable to unseen rows~\cite{han2024large,slack2023tablet}.
Notably, prior work primarily focuses on few-shot settings with a small number of labeled examples~\cite{slack2023tablet,nam2023semi,dinh2022lift,hegselmann2023tabllm}.
Recent efforts have begun exploring zero-shot tabular learning by independently prompting LLMs on individual rows~\cite{slack2023tablet,hegselmann2023tabllm}, but such approaches overlook implicit relationships between rows.
To address this limitation, we use LLMs to extract higher-order relationships among rows, even without requiring row-wise inputs.

\smallsection{LLMs for Hypergraphs.}
Recent studies have explored integrating LLMs with graphs~\cite{li2023survey,jin2024large,tang2024graphgpt,zhang2024graphtranslator} and extended these efforts to hypergraphs.
Early approaches focus on latent-level fusion, combining representations from LLMs and hypergraph encoders to capture semantic and structural information~\cite{bazaga2024hyperbert,liedullm}.
More recent studies use LLMs to reason over small hypergraphs via textualized higher-order relations~\cite{feng2024beyond}, or to generate hypergraphs that preserve structural patterns in real-world data~\cite{gu2025modeling}.
In contrast, in our work, LLMs generate hypergraphs by grouping semantically related rows, grounded in reasoning over tabular attributes rather than preserving structural patterns.

\smallsection{Hypergraphs for Tabular Data.}
Prior work has explored modeling tabular data using hypergraphs.
For instance, HyTrel~\cite{chen2023hytrel} represents each cell as a node and forms hyperedges over rows, columns, or the entire table to encode explicit structural dependencies.
Other approaches construct row-level hypergraphs by linking each entity to its top-$K$ related rows using BM25-style relevance~\cite{du2022learning} or to nearest neighbors in feature space~\cite{li2025prototype}.
However, these methods rely on heuristic similarity measures and lack semantic reasoning over attributes, which our approach leverages through LLMs.


\begin{figure*}[t]
    \centering
    \includegraphics[width=0.995\textwidth]{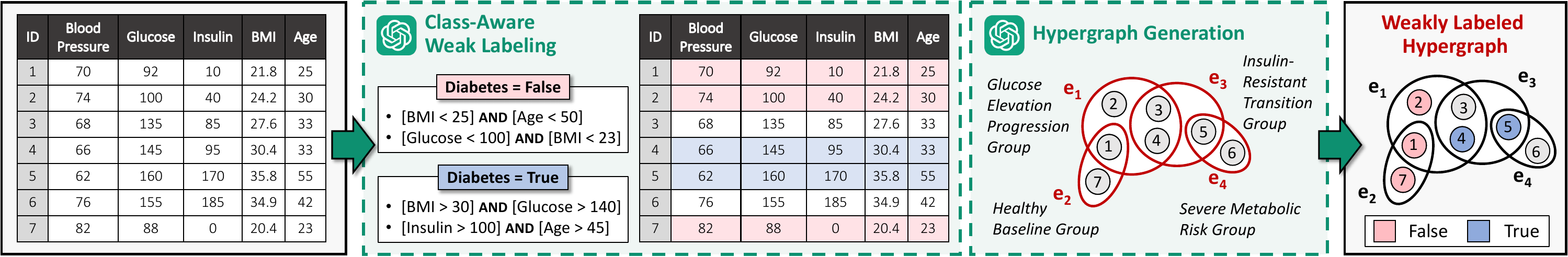} 
    \caption{
    Given an unlabeled tabular dataset, \naive employs an LLM (1) to derive attribute-based condition rules for weak label assignment and (2) to identify semantically related row groups for hypergraph construction, yielding a weakly labeled hypergraph.
    \label{fig:llm:generate}
    }
    \vspace{-5pt}
\end{figure*} 

\smallsection{Weakly-Supervised Learning.}
In the broader machine learning context beyond tabular learning,
weakly supervised learning aims to train models by leveraging and combining weak, noisy labels that are often generated by multiple rules, programs, or even LLMs ~\cite{ratner2016data, ratner2017snorkel, zhou2017brief, smith2024language}.
Prior work refines labels through various strategies, including modeling inter-label dependencies \cite{ratner2017snorkel}, accounting for source reliability \cite{ruhling2021end}, and integrating foundation model embeddings~\cite{chen2022shoring}.
The most closely related work is~\cite{rattana2023label}, which refines given weak labels via label propagation on a pairwise graph constructed using simple mechanisms, such as thresholding Euclidean distances, for datasets composed of purely numerical feature vectors.
In contrast, our approach exploits LLMs to generate both weak labels and a hypergraph through semantic reasoning over tabular attributes.



\section{Proposed Method: \method}
\label{sec:method}
\begin{figure*}[t]
    \centering
    \includegraphics[width=0.99\textwidth]{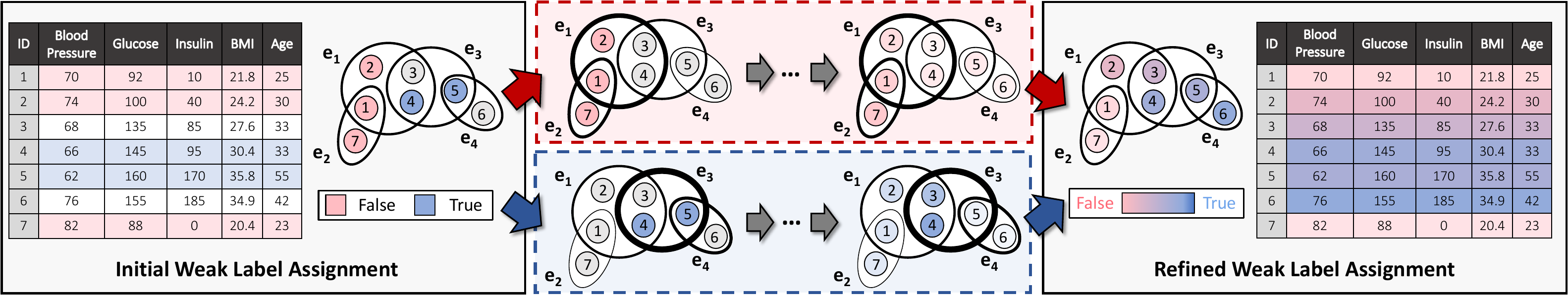} 
    \caption{
    Given the initial weakly labeled hypergraph, \method propagates each weak label independently over a label-adaptive hypergraph, where hyperedge weights (indicated by edge thickness) reflect their effectiveness for the target label.
    The resulting label-wise propagation outputs are then aggregated into refined weak label assignment probabilities, capturing higher-order inter-row dependencies.
    \label{fig:llm:tabhgr}
    }
    \vspace{-5pt}
\end{figure*}

In this section, we first define the problem (Section~\ref{sec:method:problem}).
Then, we present \method, consisting of three steps: LLM-based weak labeling (Section~\ref{sec:method:pseudo}), label refinement via hypergraph construction and propagation (Section~\ref{sec:method:naive}), and hypergraph enhancement through label-adaptive hyperedge weighting (Section~\ref{sec:method:final}).

\subsection{Problem Statement}
\label{sec:method:problem}
Let $\mathcal{D}=\{\mathbf{s}_i\}_{i=1}^{N}$ be a tabular dataset with $N$ rows and heterogeneous attributes (e.g., categorical, numerical, or ordinal), where each row $\mathbf{s}_i\in\mathcal{X}$ represents an entity. Let $d_\text{task}$ and $d_\text{attr}$ denote text descriptions of the target task and attributes, and let $\mathcal{Y}=\{1,\dots,C\}$ be the label space. 
We consider a \textit{transductive zero-shot} setting: no ground-truth labels are available, while the unlabeled rows in $\mathcal{D}$ are jointly accessible during inference.
The goal of \textit{zero-shot tabular learning} is to learn a classifier $f:\mathcal{X}\rightarrow\mathcal{Y}$ that predicts $\hat{y}_i=f(\mathbf{s}_i)$ without labeled training data.
Supervision is replaced by external semantic knowledge from off-the-shelf LLMs, which bridge the input space $\mathcal{X}$ and label space $\mathcal{Y}$.


\subsection{LLM-Based Weak Labeling}
\label{sec:method:pseudo}

In the absence of labeled data, we prompt an LLM to act as a \textit{domain expert} that characterizes each class using attribute-based conditions. The prompt includes the task description, attribute descriptions, and a small random set of unlabeled rows to ground its reasoning (see Appendix~\ref{app:prompt}).


For each class $c \in \{1, \dots, C\}$, the LLM generates $L$ interpretable \textit{condition rules} over attributes:
\begin{equation*}
\begin{aligned}
    \mathcal{R}^{(c)}
    &= \{r_{1}^{(c)}, r_{2}^{(c)}, \dots, r_{L}^{(c)}\} \\
    &= \texttt{LLM}\!\left(
    \mathcal{P}_{\text{class}}(c, L, d_\text{task}, d_\text{attr})
    \right).
\end{aligned}
\end{equation*}
where $\mathcal{P}_{\text{class}}$ denotes a prompt template that instructs the LLM to describe class $c$ by producing $L$ condition rules based on the task and attribute information. 
As illustrated in Figure~\ref{fig:llm:generate}, each rule $r_j^{(c)} \in \mathcal{R}^{(c)}$ is a conjunction of attribute–value conditions that characterize class $c$ (e.g., \texttt{\small [insulin\;$>$\;150]} and \texttt{\small [BMI\;$>$\;30]} for the class \texttt{\small diabetes\;=\;true}). 
We define the complete set of rules across all classes as $\mathcal{R} = \bigcup_{c=1}^{C} \mathcal{R}^{(c)}$ where $|\mathcal{R}|=C\cdot L$.

Then, each row $\mathbf{s}_i \in \mathcal{D}$ is evaluated against every generated rule, yielding a binary \textit{weak label}:
\begin{equation*}\label{eq:llm:condition}
x_{i,j}^{(c)} =
\begin{cases}
1, & \text{if } \mathbf{s}_i \text{ satisfies } r_{j}^{(c)},\\
0, & \text{otherwise.}
\end{cases}
\end{equation*}
This induces a weak label assignment vector associated with rule $r_j^{(c)}$:
\begin{equation}
\label{eq:init:vec}
    \mathbf{x}_{j}^{(c)} = \left[x_{1,j}^{(c)}, \dots, x_{N,j}^{(c)}\right]^\top \in \{0,1\}^{N}.
\end{equation}

We obtain class-level predictions by aggregating the weak label assignments across rules.
Specifically, the compatibility score of class $c$ for row $\mathbf{s}_i$ is computed by summing the weak labels over the $L$ rules associated with class $c$:
\begin{equation}\label{eq:z}
z_i^{(c)} = \sum\nolimits_{j=1}^{L} \left[\mathbf{x}_j^{(c)}\right]_i.
\end{equation}
If necessary, we can obtain predictions by normalizing these scores across all classes:
\begin{equation}\label{eq:y_hat}
    \hat{y}_i^{(c)} =
\frac{z_i^{(c)}}{\sum_{c'=1}^{C} z_i^{(c')}}.
\end{equation}

Importantly, weak labeling requires only a constant number of LLM prompts without feeding individual rows, making it independent of the row count and more efficient than row-wise prompting.
However, the resulting weak labels are derived from coarse rules and are therefore inherently noisy.
Moreover, they are assigned independently to rows and thus fail to capture inter-row dependencies.


\subsection{Hypergraph-Based Label Refinement}
\label{sec:method:naive}

To address the limitations of weak labels, we refine them by leveraging inter-row relations.
We first use an LLM to identify semantically related groups of rows, which induces a hypergraph. Weak labels are then propagated over this hypergraph to refine their assignment probabilities.





\smallsection{Hypergraph Construction using LLMs.}
In tabular data, rows often share attribute patterns, and jointly considering their weak labels can facilitate refinement.
To capture such dependencies, we prompt the LLM to construct attribute-based condition rules that define \textit{semantic groups}, which are subsets of related rows, as follows:
\begin{equation*}
\begin{aligned}
\mathcal{R}_\text{group}
&= \{\tilde{r}_1,\tilde{r}_2,\dots,\tilde{r}_K\} \\
&= \texttt{LLM}\!\left(
\mathcal{P}_{\text{group}}(K,d_\text{task},d_\text{attr})
\right).
\end{aligned}
\end{equation*}
where $\mathcal{P}_{\text{group}}$ denotes a prompt template that instructs the LLM to generate $K$ semantic grouping rules (see Appendix~\ref{app:prompt} for the prompt template). 
Each rule $\tilde{r}_k\in\mathcal{R}_\text{group}$ is a conjunction of attribute-value conditions that defines a row group $e_k$. For example, in Figure~\ref{fig:llm:generate}, $\tilde{r}_3$ is characterized by \texttt{\small [70\;$<$\;insulin\;$<$\;175]} and \texttt{\small [BMI\;$>$\;27]}. The rows satisfying this rule form the semantic group $e_3$, referred to as the \textit{insulin-resistant transition group}. Note that row grouping requires only a few examples in LLM prompts, rather than all rows.


These groups define a hypergraph $\mathcal{G}=(\mathcal{V},\mathcal{E})$, where nodes are rows ($|\mathcal{V}|=N$) and hyperedges are LLM-induced semantic groups.


\smallsection{Hypergraph-based Label Propagation.}
Given the constructed hypergraph, \naive refines weak label assignments by propagating them over the hypergraph.
Let $H\in\{0,1\}^{|\mathcal{V}|\times|\mathcal{E}|}$ denote the \textit{incidence matrix} of $\mathcal{G}$, where $H_{i,k}=1$ if row $\mathbf{s}_i$ belongs to group $e_k$.
The node and hyperedge degree matrices are defined as
$D_V = \mathrm{diag}(H \mathbf{1}) \in \mathbb{R}^{|\mathcal{V}|\times|\mathcal{V}|}$
and
$D_E = \mathrm{diag}(H^\top \mathbf{1}) \in \mathbb{R}^{|\mathcal{E}|\times|\mathcal{E}|}$,
respectively.

For each rule $r_j^{(c)}\in\mathcal{R}$, we use its initial weak label assignment vector $\mathbf{x}_j^{(c,0)}=\mathbf{x}_j^{(c)}$ (Eq.~\eqref{eq:init:vec}) as the initialization for propagation.
These weak labels are iteratively propagated over the hypergraph as:
\begin{equation*}
\mathbf{x}_{j}^{(c,t+1)} =
\alpha \, D_V^{-1} H D_E^{-1} H^\top \mathbf{x}_{j}^{(c,t)}
+ (1-\alpha)\,\mathbf{x}_{j}^{(c,0)},
\end{equation*}
where $\mathbf{x}_{j}^{(c,t)} \in \mathbb{R}^{|\mathcal{V}|}$ denotes the propagated weak label probability vector at the $t$\textsuperscript{th} iteration, and $\alpha \in [0,1]$ controls the propagation strength.
After $T$ iterations, we obtain the refined vector $\mathbf{x}_{j}^{(c,T)} \in \mathbb{R}^{|\mathcal{V}|}$.
Note that this process requires only a forward pass on the hypergraph, without any training.

For prediction, we compute class-level compatibility scores using the refined labels by replacing the initial assignments $\mathbf{x}_j^{(c,0)}$ ($=\mathbf{x}_j^{(c)}$) with the propagated vectors $\mathbf{x}_j^{(c,T)}$ in Eq.~\eqref{eq:z}.

In essence, label propagation refines each row’s weak labels by aggregating labels from related rows while accounting for higher-order relationships captured by the hypergraph. This improves zero-shot performance, as shown in Section~\ref{sec:exp}.


\subsection{Label-Adaptive Hyperedge Weighting}
\label{sec:method:final}

Although LLM-generated hypergraphs use semantic attribute reasoning, they may be suboptimal for propagation. We therefore enhance them with label-adaptive hyperedge weights.




\smallsection{Analysis of Label–Structure Dependency.}
We first analyze how each hyperedge contributes to weak label propagation. Intuitively, a hyperedge may help propagate some weak labels while being ineffective or harmful for others.

To quantify this effect, we define \textit{hyperedge effectiveness} by measuring the change in prediction quality when a hyperedge is removed.
Specifically, for a hyperedge $e_k \in \mathcal{E}$ and a weak-label rule $r_j^{(c)} \in \mathcal{R}$, we define:
\begin{equation}\label{eq:eff_score}
\scalebox{0.82}{$
\Delta(e_k \mid r_j^{(c)}) =
\mathcal{M}\!\left(\mathbf{x}_{j}^{(c,T)}(\mathcal{G}),\, Y\right)
-
\mathcal{M}\!\left(\mathbf{x}_{j}^{(c,T)}(\mathcal{G}_{\setminus e_k}),\, Y\right),
$}
\end{equation}
where $\mathbf{x}_{j}^{(c,T)}(\mathcal{G})$ denotes the propagated weak label scores obtained on hypergraph $\mathcal{G}$ under rule $r_j^{(c)}$, and $\mathcal{G}_{\setminus e_k}$ is the hypergraph with hyperedge $e_k$ removed.
Here, $\mathcal{M}(\cdot,\cdot)$ is a scalar performance metric instantiated as AUROC and evaluated against ground-truth labels $Y$.
This definition is used only for analysis to study how weak label semantics depend on hypergraph structure.~\footnote{Since ground-truth labels are unavailable in the zero-shot setting, this score is later estimated in an unsupervised manner.} A positive $\Delta(e_k \mid r_j^{(c)})$ indicates that $e_k$ supports propagation for rule $r_j^{(c)}$, while a negative value indicates an adverse effect.

\begin{figure}[t]
    \centering
    \begin{subfigure}{0.46\linewidth}
        \centering
        \includegraphics[width=\linewidth]{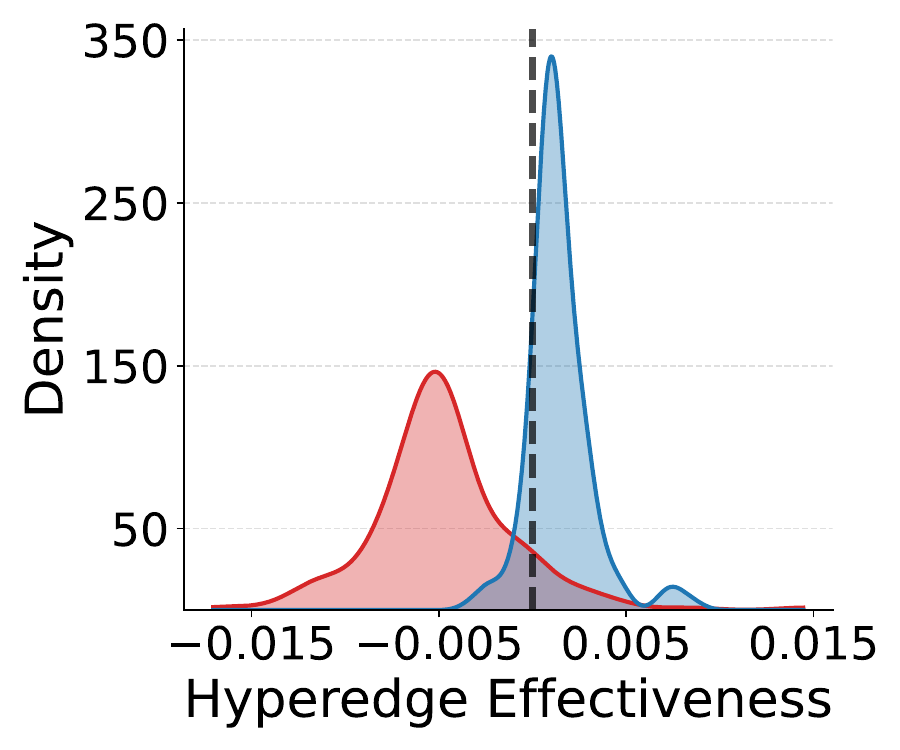}
        \caption{Effectiveness density}
        \label{fig:llm:eff:a}
    \end{subfigure}
    \hspace{5pt}
    \begin{subfigure}{0.46\linewidth}
        \centering
        \includegraphics[width=\linewidth]{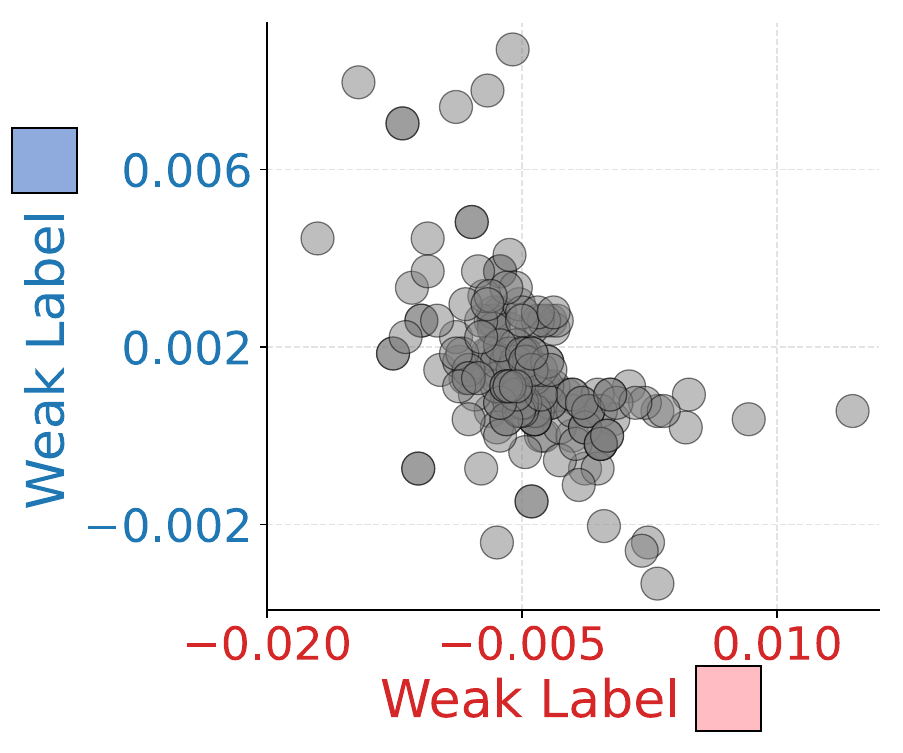}
        \caption{Correlation btw. rules}
        \label{fig:llm:eff:b}
    \end{subfigure}
    \caption{
    The distribution of hyperedge effectiveness varies across weak labels (\textcolor{PseudoFalse}{\(\blacksquare\)} and \textcolor{PseudoTrue}{\(\blacksquare\)}).
    Moreover, hyperedge effectiveness with respect to weak labels is not necessarily positively correlated and may be weakly correlated or even negatively correlated.
    }
    \label{fig:llm:eff}
    \vspace{-5pt}
\end{figure}

As shown in Figure~\ref{fig:llm:eff:a}, hyperedge effectiveness varies across weak labels, and Figure~\ref{fig:llm:eff:b} shows that these patterns can be weakly or even negatively correlated. Thus, hyperedges useful for one weak label may be ineffective or harmful for others, making the constructed hypergraph suboptimal for universal weak-label propagation.

Motivated by this analysis, we introduce label-adaptive hyperedge weights, enabling propagation to focus on effective hyperedges. These weights are estimated without ground-truth labels.

\smallsection{Hyperedge Effectiveness Estimation.}
In the zero-shot setting, ground-truth labels $Y$ are unavailable, making direct evaluation of hyperedge effectiveness (Eq.~\eqref{eq:eff_score}) infeasible.
To address this, we first estimate hyperedge effectiveness in an \textit{unsupervised} manner by using the predicted labels:
\begin{equation*}
\widehat{Y}=\{\hat{c}_i\}_{i=1}^N,\;\;\;\;\hat{c}_i = \arg\max_{c\in\{1,\cdots,C\}} \hat{y}_i^{(c)},
\end{equation*}
using Eq.~\eqref{eq:y_hat} as a proxy for ground-truth labels.

\begin{figure}[t]
    \centering
    \begin{subfigure}{0.45\linewidth}
        \centering
        \includegraphics[width=\linewidth]{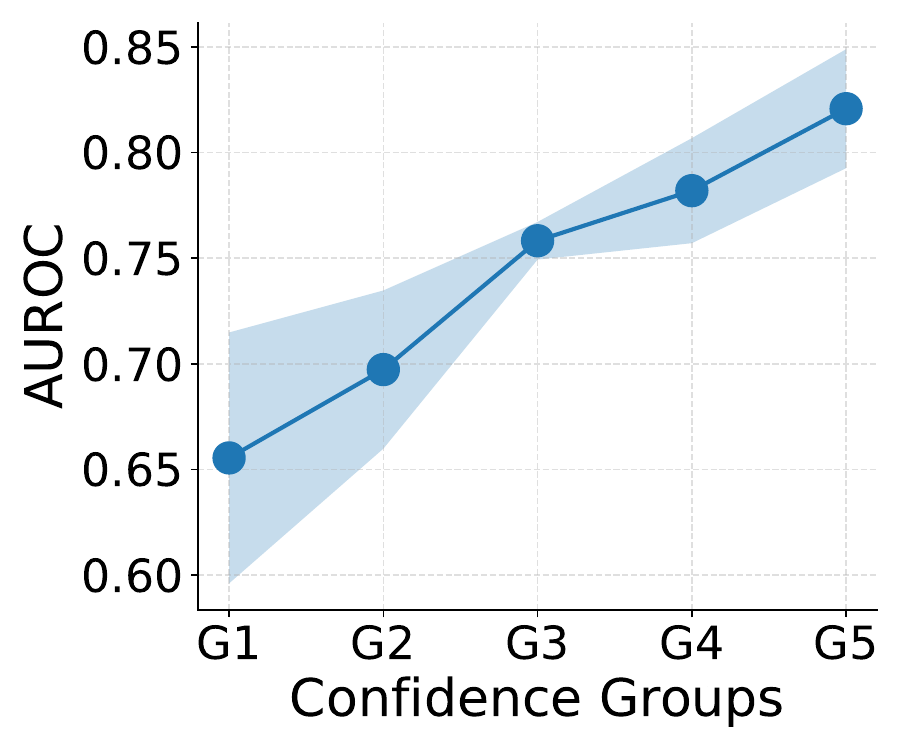}
        \caption{Diabetes}
        \label{fig:conf:a}
    \end{subfigure}
    \hspace{5pt}
    \begin{subfigure}{0.45\linewidth}
        \centering
        \includegraphics[width=\linewidth]{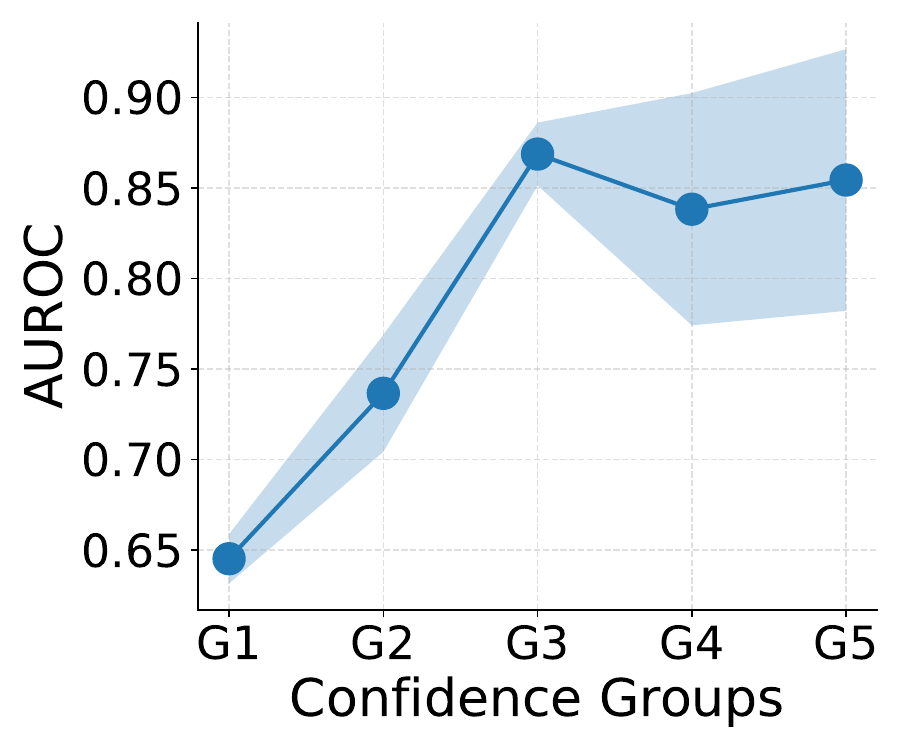}
        \caption{Heart}
        \label{fig:conf:b}
    \end{subfigure}
    \caption{
    Rows are grouped from G1 to G5 by confidence, defined as the margin between the top two compatibility scores (Eq.~\eqref{eq:z}). Since propagation is more accurate on high-confidence rows, we use them to estimate hyperedge effectiveness.
    }
    \label{fig:conf}
    \vspace{-5pt}
\end{figure}

To ensure reliability, we define each row’s confidence as the margin between its top two compatibility scores (Eq.~\eqref{eq:z}), i.e., $\max_c z_i^{(c)} - \max_{c' \neq c} z_i^{(c')}$. A larger margin indicates stronger class separation. 
Then, we retain the top-$q$\% most confident rows, forming $\mathcal{D}_q\subseteq\mathcal{D}$ with weak labels $\widehat{Y}_q$.

As shown in Figure~\ref{fig:conf}, hypergraph-based propagation (Section~\ref{sec:method:naive}) is more accurate on high-confidence rows. We therefore use them as reliable proxies for estimating hyperedge effectiveness:
\begin{equation}\label{eq:eff_score:estimated}
\scalebox{0.832}{$
\widetilde{\Delta}(e_k \mid r_j^{(c)}) =
\mathcal{M}\!\left(\mathbf{x}_{j}^{(c,T)}(\mathcal{G}),\, \widehat{Y}_q\right)
-
\mathcal{M}\!\left(\mathbf{x}_{j}^{(c,T)}(\mathcal{G}_{\setminus e_k}),\, \widehat{Y}_q\right),
$}
\end{equation}
which replaces the ground-truth labels $Y$ in Eq.~\eqref{eq:eff_score} with the proxy labels $\widehat{Y}_q$ from confident rows $\mathcal{D}_q$.

\smallsection{Label-adaptive Hyperedge Weighting.}
The estimated effectiveness of each hyperedge for weak label rule $r_j^{(c)}$ (Eq.~\eqref{eq:eff_score:estimated}) is used as its label-specific weight, $w_{k,j}^{(c)}$.~\footnote{To obtain non-negative weights, we apply a softmax:
{\scriptsize
\begin{equation*}
w_{k,j}^{(c)}
=
|\mathcal{E}|
\frac{
\exp\!\left(\widetilde{\Delta}(e_k\mid r_j^{(c)})/\tau\right)
}{
\sum_{\ell=1}^{|\mathcal{E}|}
\exp\!\left(\widetilde{\Delta}(e_\ell\mid r_j^{(c)})/\tau\right)
},
\end{equation*}
}
where $\tau$ controls the concentration.}
Then, we define the hyperedge weight vector as:
\begin{equation*}
    \mathbf{w}_{j}^{(c)} = \left[w_{1,j}^{(c)}, w_{2,j}^{(c)}, \dots, w_{|\mathcal{E}|,j}^{(c)}\right]^\top \in \mathbb{R}^{|\mathcal{E}|}.
\end{equation*}
By Eq.~\eqref{eq:eff_score:estimated}, $w_{k,j}^{(c)}$ measures how much hyperedge $e_k$ contributes to propagating weak labels from $r_j^{(c)}$.

Using these label-specific weights, propagation for each weak label is performed as:
\begin{equation*}
\scalebox{0.87}{$
\mathbf{x}_{j}^{(c,t+1)} = \alpha \; D_V^{-1} H \mathbf{W}_{j}^{(c)} D_E^{-1} H^\top \mathbf{x}_{j}^{(c,t)} + (1-\alpha)\; \mathbf{x}_{j}^{(c,0)},
$}
\end{equation*}
where $\mathbf{W}_{j}^{(c)} = \mathrm{diag}(\mathbf{w}_{j}^{(c)})\in\mathbb{R}^{|\mathcal{E}|\times |\mathcal{E}|}$ encodes the weak label-specific relevance of hyperedges.
As before, this process requires only a forward pass and involves no training.

For prediction, class-level compatibility scores are computed from the weak labels refined by label-adaptive propagation by substituting the initial assignments $\mathbf{x}_j^{(c,0)}$ ($=\mathbf{x}_j^{(c)}$) in Eq.~\eqref{eq:z} with the propagated vectors $\mathbf{x}_j^{(c,T)}$.


This enables hypergraph propagation to prioritize hyperedges that are more effective for each weak label. We empirically validate this label-adaptive propagation in Section~\ref{sec:exp}. 

\smallsection{Complexity.}
\method requires $\lceil|\mathcal{R}|/B\rceil+\lceil|\mathcal{E}|/B\rceil$ LLM calls, independent of the number of rows, and has time complexity $O(|\mathcal{R}|T|\mathcal{E}|\sum_{e\in\mathcal{E}}|e|)$ due to hyperedge effectiveness estimation. 
Refer to Appendix~\ref{appendix:complexity} for further details.

\section{Experimental Results}
\label{sec:exp}

We evaluate the accuracy and effectiveness of \method by answering the research questions:
\begin{itemize}[leftmargin=*, noitemsep, topsep=0pt]
\item \textbf{Q1. Accuracy.} How accurate is \method compared to existing zero/few-shot baselines?
\item \textbf{Q2. Efficiency.} How efficient is \method in terms of LLM usage and inference cost?
\item \textbf{Q3. Effectiveness.} Do LLM-generated hypergraphs improve zero-shot prediction?
\item \textbf{Q4. Inductive Applicability.}
Can \method generalize to unseen rows?
\item \textbf{Q5. Parameter Analysis.} How do the key parameters of \method affect performance?
\end{itemize}

\begin{table}[t]
\centering
\caption{Statistics of the datasets.
Cat.\ and Num.\ denote the numbers of categorical and numerical attributes.}
\label{tab:llm:dataset}
\scalebox{0.77}{
\begin{tabular}{lcccc}
\toprule
\textbf{Data} & \textbf{\# of Rows} & \textbf{Cat.} & \textbf{Num.} & \textbf{Label ratio (\%)} \\
\midrule
Adult          & 48,842 & 7   & 7   & 76:24 \\
Bank           & 45,211 & 8   & 8   & 88:12 \\
Blood          & 748    & 0   & 4   & 76:24 \\
Car            & 1,728  & 5   & 1   & 70:22:4:4 \\
Credit-G       & 1,000  & 12  & 8   & 70:30 \\
Diabetes       & 768    & 0   & 8   & 65:35 \\
Heart          & 918    & 4   & 7   & 45:55 \\
Communities    & 1,994  & 1   & 102 & 34:33:33 \\
Myocardial     & 1,700  & 94  & 17  & 22:78 \\
\bottomrule
\end{tabular}}
\end{table}

\begin{table*}[t!]
\begin{center}
\caption{\label{tab:llm:results} 
\underline{\smash{Overall performance comparison.}}  
\method outperforms existing zero-shot and 4-, 8-, and 16-shot tabular learning methods in AUROC across nine benchmarks, achieving the best average rank. Best results are in \textbf{bold}, second-best are \underline{\smash{underlined}}, and ranks are computed within each shot setting.
For methods with missing results, the average rank is computed only over datasets with valid results.
}
\setlength\tabcolsep{4.8pt} 
\scalebox{0.735}{
\begin{tabular}{cc|ccccccccc|c}
    \toprule
     Method & Shot & Adult & Bank & Blood & Car & Credit-G & Diabetes & Heart & \makecell[c]{Comm-\\unities} & \makecell[c]{Myoca-\\rdial} & Avg. Rank \\
    \midrule
    \multirow{3}{*}{LogReg} 
    & 4 & 0.7210 & 0.6370 & 0.5679 & 0.6238 & 0.5268 & 0.5709 & 0.7054 & 0.6745 & 0.5125 & 7.00\\
    & 8 & 0.7602 & 0.7252 & 0.6851 & 0.7205 & 0.5552 & 0.6552 & 0.7812 & 0.7373 & 0.5534 & 5.89\\
    & 16 & 0.7520 & \underline{\smash{0.7751}} & 0.6830 & 0.8242 & 0.5826 & 0.7344 & 0.8302 & 0.7255 & 0.6000 & 5.22\\
    \midrule
    \multirow{3}{*}{\makecell[c]{XGBoost\\\cite{chen2016xgboost}}}
    & 4 & 0.5000 & 0.5000 & 0.5000 & 0.5000 & 0.5000 & 0.5000 & 0.5000 & 0.5000 & 0.5394 & 10.33\\
    & 8 & 0.5919 & 0.5878 & 0.5997 & 0.6400 & 0.5222 & 0.5086 & 0.5588 & 0.6665 & 0.5563 & 10.33\\
    & 16 & 0.6068 & 0.7034 & 0.6328 & 0.7226 & 0.5623 & 0.6559 & 0.7862 & 0.6801 & 0.5655 & 9.00\\
    \midrule
    \multirow{3}{*}{\makecell[c]{SCARF\\\cite{bahri2021scarf}}}
    & 4 & 0.5834 & 0.5853 & 0.5622 & 0.6252 & 0.4892 & 0.6235 & 0.5938 & 0.6618 & 0.4770 & 9.00\\
    & 8 & 0.7242 & 0.5528 & 0.6577 & 0.7223 & 0.5526 & 0.6469 & 0.7435 & 0.7269 & 0.4937 & 8.11\\
    & 16 & 0.7563 & 0.6581 & 0.6627 & 0.7577 & \underline{\smash{0.5922}} & 0.7186 & 0.8366 & 0.7309 & 0.5431 & 7.11\\
    \midrule
    \multirow{3}{*}{\makecell[c]{TabPFN\\\cite{hollmann2022tabpfn}}}
    & 4 & 0.6089 & 0.6319 & 0.5872 & 0.5814 & 0.5400 & 0.5628 & 0.6733 & - & - & 8.00\\
    & 8 & 0.7042 & 0.6281 & 0.6630 & 0.6395 & 0.5258 & 0.6908 & 0.7789 & - & - & 8.86\\
    & 16 & 0.7034 & 0.7379 & 0.6414 & 0.7135 & 0.5891 & 0.7369 & 0.8145 & - & - & 7.43\\
    \midrule
    \multirow{3}{*}{\makecell[c]{STUNT\\\cite{nam2023stunt}}}
    & 4 & 0.6743 & 0.5634 & 0.4857 & 0.6132 & 0.4880 & 0.6422 & 0.8827 & 0.6687 & 0.5277 & 8.33 \\
    & 8 & 0.8216 & 0.6301 & 0.6000 & 0.6786 & 0.5450 & 0.6739 & 0.8878 & 0.7636 & 0.5540 & 6.67\\
    & 16 & 0.8057 & 0.6985 & 0.5476 & 0.7556 & 0.5763 & 0.7379 & 0.8913 & \underline{\smash{0.7729}} & \underline{\smash{0.6122}} & 5.78\\
    \midrule
    \multirow{3}{*}{\makecell[c]{In-Context\\\cite{wei2022emergent}}}
    & 4 & 0.7751 & 0.6138 & 0.5630 & 0.6247 & 0.5299 & 0.7171 & 0.6076 & - & - & 7.00 \\
    & 8 & 0.7930 & 0.6957 & 0.5899 & 0.6757 & 0.5243 & 0.7221 & 0.6546 & - & - & 8.57 \\
    & 16 & 0.7950 & 0.6976 & 0.5659 & 0.7694 & 0.5529 & 0.7164 & 0.6700 & - & - & 9.14 \\
    \midrule
    \multirow{4}{*}{\makecell[c]{TABLET\\\cite{slack2023tablet}}}
    & 0 & 0.7723 & 0.6910 & 0.5139 & 0.7785 & 0.4681 & 0.6946 & 0.7916 & 0.6604 & 0.6103 & 4.33\\
    & 4 & 0.7529 & 0.5811 & 0.5645 & 0.6021 & 0.5433 & 0.6396 & 0.6819 & - & - & 7.57 \\
    & 8 & 0.7756 & 0.6908 & 0.5637 & 0.6553 & 0.5290 & 0.6547 & 0.6985 & - & - & 9.14 \\
    & 16 & 0.7974 & 0.6940 & 0.6062 & 0.7402 & 0.5462 & 0.6671 & 0.6839 & - & - & 10.00 \\
    \midrule
    \multirow{4}{*}{\makecell[c]{TabLLM\\\cite{hegselmann2023tabllm}}}
    & 0 & 0.7969 & 0.6220 & 0.5392 & 0.7812 & 0.5236 & 0.6940 & 0.7824 & 0.6809 & 0.5818 & 4.33\\
    & 4 & 0.8357 & 0.6251 & 0.5587 & 0.8582 & 0.5190 & 0.7042 & 0.5974 & - & - & 6.86\\
    & 8 & 0.8352 & 0.6319 & 0.6601 & \underline{\smash{0.8743}} & 0.5642 & 0.6430 & 0.7014 & - & - & 6.14\\
    & 16 & 0.8323 & 0.6373 & 0.6514 & \textbf{0.8865} & \textbf{0.6038} & 0.6734 & 0.8172 & - & - & 6.29\\
    \midrule
    \multirow{3}{*}{\makecell[c]{FeatLLM\\\cite{han2024large}}}
    & 4 & 0.8668 & 0.7045 & 0.6834 & 0.7269 & 0.5594 & \textbf{0.8028} & 0.7566 & 0.7539 & 0.5287 &  3.56 \\
    & 8 & 0.8789 & 0.7585 & 0.7037 & 0.7326 & 0.5742 & 0.7938 & 0.7946 & 0.7659 & 0.5622 & 3.11 \\
    & 16 & 0.8754 & \textbf{0.7841} & 0.7007 & 0.7943 & 0.5660 & \underline{\smash{0.8015}} & 0.8371 & 0.7625 & 0.5532 & \underline{\smash{4.00}}\\
    \midrule
    \midrule
    \multirow{4}{*}{\textbf{Weak Labels Only}} & \multirow{4}{*}{0} 
    & \multirow{4}{*}{\makecell[c]{0.8722\\{\small $\pm$0.0069}}} 
    & \multirow{4}{*}{\makecell[c]{0.6998\\{\small $\pm$0.0173}}} 
    & \multirow{4}{*}{\makecell[c]{0.7447\\{\small $\pm$0.0335}}} 
    & \multirow{4}{*}{\makecell[c]{0.7799\\{\small $\pm$0.0313}}} 
    & \multirow{4}{*}{\makecell[c]{0.5524\\{\small $\pm$0.0222}}} 
    & \multirow{4}{*}{\makecell[c]{0.7406\\{\small $\pm$0.0193}}} 
    & \multirow{4}{*}{\makecell[c]{0.8782\\{\small $\pm$0.0210}}} 
    & \multirow{4}{*}{\makecell[c]{0.7700\\{\small $\pm$0.0020}}} 
    & \multirow{4}{*}{\makecell[c]{0.5856\\{\small $\pm$0.0151}}} 
    & 3.00 \small(vs. 0-shot)\\
    &  &  &  &  &  &  &  &  &  &  & 3.33 {\small(vs. 4-shot)} \\
    &  &  &  &  &  &  &  &  &  &  & 3.89 {\small(vs. 8-shot)} \\
    &  &  &  &  &  &  &  &  &  &  & 5.11 {\small(vs. 16-shot)} \\
    \midrule
    \multirow{4}{*}{\textbf{\naive}} & \multirow{4}{*}{0} 
    & \multirow{4}{*}{\makecell[c]{\underline{\smash{0.8817}}\\{\small $\pm$0.0046}}} 
    & \multirow{4}{*}{\makecell[c]{0.7092\\{\small $\pm$0.0157}}} 
    & \multirow{4}{*}{\makecell[c]{\underline{\smash{0.7471}}\\{\small $\pm$0.0286}}} 
    & \multirow{4}{*}{\makecell[c]{0.7900\\{\small $\pm$0.0295}}} 
    & \multirow{4}{*}{\makecell[c]{0.5464\\{\small $\pm$0.0195}}} 
    & \multirow{4}{*}{\makecell[c]{0.7511\\{\small $\pm$0.0218}}} 
    & \multirow{4}{*}{\makecell[c]{\underline{\smash{0.9002}}\\{\small $\pm$0.0131}}} 
    & \multirow{4}{*}{\makecell[c]{0.7727\\{\small $\pm$0.0015}}} 
    & \multirow{4}{*}{\makecell[c]{0.5867\\{\small $\pm$0.0093}}} 
    & \underline{\smash{2.11}} \small(vs. 0-shot)\\
    &  &  &  &  &  &  &  &  &  &  & \underline{\smash{2.33}} {\small(vs. 4-shot)} \\
    &  &  &  &  &  &  &  &  &  &  & \underline{\smash{2.89}} {\small(vs. 8-shot)} \\
    &  &  &  &  &  &  &  &  &  &  & \underline{\smash{4.00}} {\small(vs. 16-shot)} \\
    \midrule
    \multirow{4}{*}{\textbf{\method}} & \multirow{4}{*}{0} 
    & \multirow{4}{*}{\makecell[c]{\textbf{0.8820}\\{\small $\pm$0.0044}}} 
    & \multirow{4}{*}{\makecell[c]{0.7202\\{\small $\pm$0.0112}}} 
    & \multirow{4}{*}{\makecell[c]{\textbf{0.7572}\\{\small $\pm$0.0440}}} 
    & \multirow{4}{*}{\makecell[c]{0.7966\\{\small $\pm$0.0297}}} 
    & \multirow{4}{*}{\makecell[c]{0.5285\\{\small $\pm$0.0182}}} 
    & \multirow{4}{*}{\makecell[c]{0.7716\\{\small $\pm$0.0174}}} 
    & \multirow{4}{*}{\makecell[c]{\textbf{0.9114}\\{\small $\pm$0.0101}}} 
    & \multirow{4}{*}{\makecell[c]{\textbf{0.7778}\\{\small $\pm$0.0018}}} 
    & \multirow{4}{*}{\makecell[c]{\textbf{0.6131}\\{\small $\pm$0.0201}}} 
    & \textbf{1.22} \small(vs. 0-shot)\\
    &  &  &  &  &  &  &  &  &  &  & \textbf{1.89} {\small(vs. 4-shot)} \\
    &  &  &  &  &  &  &  &  &  &  & \textbf{2.33} {\small(vs. 8-shot)} \\
    &  &  &  &  &  &  &  &  &  &  & \textbf{2.89} {\small(vs. 16-shot)} \\
    \bottomrule
\end{tabular}}
\end{center}
\end{table*}

\begin{figure*}[t]
    \centering
    \includegraphics[width=0.335\textwidth]{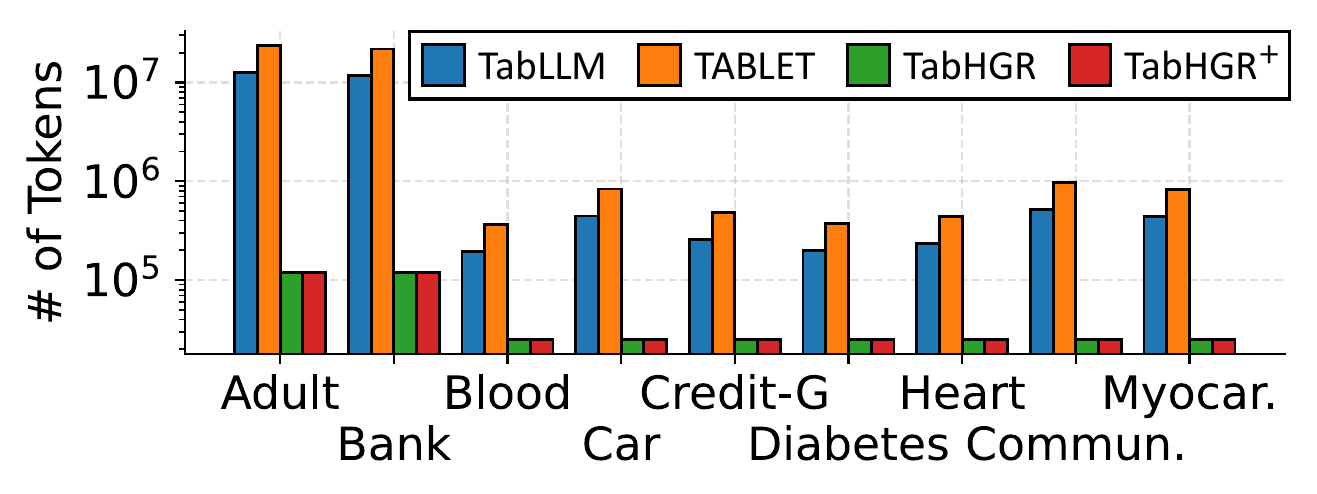}
    \hspace{-8pt}
    \includegraphics[width=0.335\textwidth]{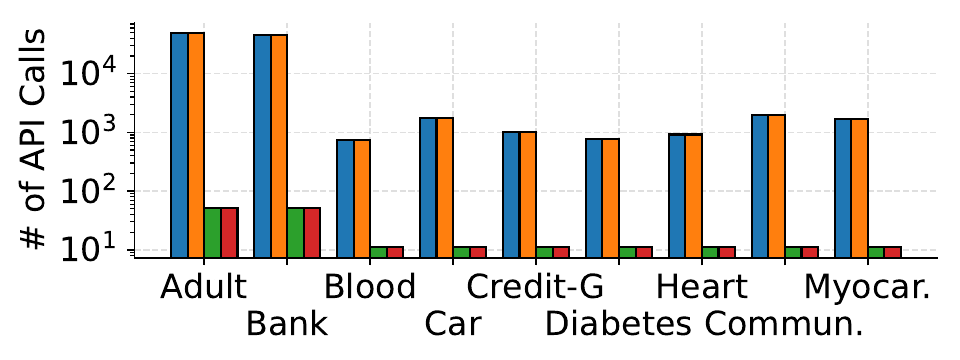}
    \hspace{-8pt}
    \includegraphics[width=0.335\textwidth]{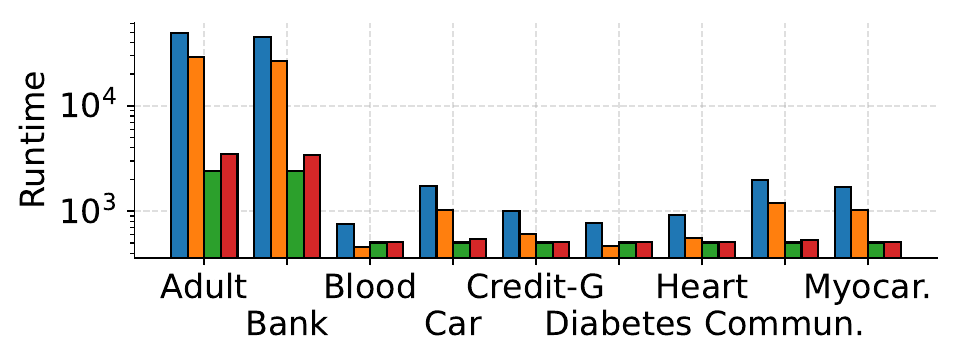} 
    \caption{
    \underline{\smash{Efficiency analysis.}}
    \method and its simple variant (\naive) are substantially more efficient than existing row-wise zero-shot LLM methods. 
    While row-wise approaches incur inference costs that scale with the row count, \method and \naive require only a constant number of input/output tokens and LLM calls, independent of dataset size. 
    Hypergraph propagation and hyperedge effectiveness estimation are performed without extra LLM queries and incur negligible overhead compared to LLM API calls.
    \label{fig:efficiency}
    }
\end{figure*}

\subsection{Experimental Settings}
We first describe the experimental settings.

\smallsection{Datasets.}
We evaluate our methods on nine publicly available tabular datasets covering diverse domains, including income prediction (Adult~\cite{asuncion2007uci}), marketing (Bank~\cite{moro2014data}), healthcare (Blood~\cite{yeh2009knowledge}, Diabetes~\footnote{\tiny\url{kaggle.com/datasets/uciml/pima-indians-diabetes-database}}, Heart~\footnote{\tiny\url{kaggle.com/datasets/fedesoriano/heart-failure-prediction}}, Myocardial~\cite{golovenkin2020myocardial}), credit assessment (Credit-G~\cite{kadra2021well}), automobile evaluation (Car~\cite{kadra2021well}), and community-level crime modeling (Communities~\cite{redmond2009communities}).
Refer to Table~\ref{tab:llm:dataset} for the statistics and Appendix~\ref{app:exp_settings} for further details.


\smallsection{Baselines.}
We compare our framework with nine representative baselines spanning traditional and recent LLM-based methods: Logistic Regression (LogReg), XGBoost~\cite{chen2016xgboost}, SCARF~\cite{bahri2021scarf}, TabPFN~\cite{hollmann2022tabpfn}, STUNT~\cite{nam2023stunt}, In-Context Learning~\cite{wei2022emergent}, TABLET~\cite{slack2023tablet}, TabLLM~\cite{hegselmann2023tabllm}, and FeatLLM~\cite{han2024large}. Most require few-shot supervision or labeled data, while TABLET and TabLLM support row-wise zero-shot prediction. We also evaluate two variants of \method: Weak Labels Only, which uses weak labels without propagation, and \naive, which removes label-adaptive hyperedge weighting.

\smallsection{Implementation.}
Note that hyperparameter tuning is prohibited in our zero-shot setting; thus, all parameters are fixed across datasets. 
We use GPT-4o~\cite{hurst2024gpt} as the default LLM and generate 20 condition rules per class for weak labeling.
By default, we generate 200 semantic groups (hyperedges) per dataset, and 1000 for larger datasets such as Adult and Bank to improve row coverage. 
Due to LLM output-length limits, we generate semantic groups through multiple independent calls, each producing 20 groups from a newly sampled set of five unlabeled rows. 
We aggregate the groups generated across all calls to construct the final hypergraph, repeating this process until reaching the target number of groups.
We set the propagation coefficient to $\alpha=0.7$, propagation steps to 2, and confidence ratio to $q=50$ by default. 
We conduct five independent runs and report the average performance of \method and its baselines.


\subsection{Q1. Overall Accuracy}

We first evaluate the prediction accuracy of \method. As shown in Table~\ref{tab:llm:results}, our zero-shot approaches, including weak labeling, \naive, and \method, substantially outperform row-wise zero-shot LLM baselines, TABLET and TabLLM. This suggests that class-aware LLM-generated rules provide more reliable semantic supervision than row-wise prompting. Moreover, despite being fully zero-shot, \naive and \method outperform strong few-shot baselines; for instance, \method exceeds 16-shot baselines on Adult, Blood, and Heart, and achieves the best average rank across 4-, 8-, and 16-shot settings. These results show the effectiveness of leveraging inter-row relationships via hypergraph generation and label propagation for zero-shot tabular learning.

\subsection{Q2. Efficiency Analysis}\label{sec:exp:efficiency}
We compare the efficiency of \method and \naive with row-wise LLM baselines in terms of tokens, API calls, and runtime. As shown in Figure~\ref{fig:efficiency}, row-wise methods scale with the number of rows since each row is queried by the LLM. In contrast, \method and \naive require only a constant number of LLM calls, independent of dataset size. Subsequent hypergraph propagation and effectiveness estimation require no additional LLM queries and add negligible overhead. Thus, our methods achieve much lower inference cost while improving predictive performance.

\subsection{Q3. Effectiveness Analysis}
We evaluate the effectiveness of hypergraphs generated by LLMs in \method and \naive.

\begin{table}[t]
\begin{center}
\caption{\label{tab:llm:random_hypergraph} 
\underline{\smash{Effectiveness of LLM-generated hypergraphs.}}
\method and its simple variant (\naive) with hypergraphs generated by LLMs outperform their kNN-based and randomized counterparts, showing the effectiveness of LLMs in inducing meaningful inter-row relations from tabular attributes in a zero-shot setting. 
}
\setlength\tabcolsep{3.6pt} 
\scalebox{0.705}{
\begin{tabular}{c|c|ccc|ccc}
    \toprule
    \multirow{2}{*}{Dataset} & \multirow{2}{*}{Raw} & \multicolumn{3}{c|}{\naive} & \multicolumn{3}{c}{\method} \\
    &  & \cellcolor{gray!15}LLM & KNN & Rand. & \cellcolor{gray!15}LLM & KNN & Rand. \\
    \midrule
    Adult 
    & 0.8722 
    & \cellcolor{gray!15}\underline{\smash{0.8817}} & 0.8429 & 0.8627 
    & \cellcolor{gray!15}\textbf{0.8820} & 0.8530 & 0.8650\\
    Bank 
    & 0.6998 
    & \cellcolor{gray!15}\smash{0.7092} & 0.7088 & 0.6906 
    & \cellcolor{gray!15}\textbf{0.7202} & \underline{0.7155} & 0.6895\\
    Blood 
    & 0.7447
    & \cellcolor{gray!15}\underline{\smash{0.7471}} & 0.7382 & 0.7342 
    & \cellcolor{gray!15}\textbf{0.7572} & 0.7408 & 0.7329\\
    Car
    & 0.7799
    & \cellcolor{gray!15}\underline{\smash{0.7900}} & 0.7882 & 0.7830
    & \cellcolor{gray!15}\textbf{0.7966} & 0.7815 & 0.7814\\
    Credit-G 
    & \smash{0.5524}
    & \cellcolor{gray!15}0.5464 & \textbf{0.5658} & 0.5513
    & \cellcolor{gray!15}0.5285 & 0.5587 & \underline{0.5644}\\
    Diabetes
    & 0.7406 
    & \cellcolor{gray!15}\underline{\smash{0.7511}} & 0.7125 & 0.7459 
    & \cellcolor{gray!15}\textbf{0.7716} & 0.7086 & 0.7351\\
    Heart 
    & 0.8782
    & \cellcolor{gray!15}\underline{\smash{0.9002}} & 0.8238 & 0.8657 
    & \cellcolor{gray!15}\textbf{0.9114} & 0.8141 & 0.8690\\
    Commu.
    & 0.7700 
    & \cellcolor{gray!15}\smash{0.7727} & \textbf{0.7809} & 0.7645
    & \cellcolor{gray!15}0.7778 & \underline{0.7793} & 0.7651\\
    Myocar.
    & 0.5856 
    & \cellcolor{gray!15}\underline{\smash{0.5867}} & 0.5839 & 0.5770
    & \cellcolor{gray!15}\textbf{0.6131} & 0.5579 & 0.5489\\
    \midrule
    \multirow{2}{*}{Avg.}
    & 0.7359 
    & \cellcolor{gray!15}\underline{\smash{0.7428}} & 0.7272 & 0.7305 
    & \cellcolor{gray!15}\textbf{0.7509} & 0.7256 & 0.7279 \\
    & (4.11)
    & \cellcolor{gray!15}\underline{\smash{(2.67)}} & (4.11) & (5.22) 
    & \cellcolor{gray!15}\textbf{(1.89)} & (4.56) & (5.33) \\
    \bottomrule
\end{tabular}}
\end{center}
\end{table}

\smallsection{Alternative Hypergraphs.}
We compare LLM-generated hypergraphs with kNN hypergraphs based on raw attribute similarity and randomized hypergraphs preserving node degrees and hyperedge sizes~\cite{lee2021hyperedges}.
Table~\ref{tab:llm:random_hypergraph} shows that \method and \naive perform best with LLM-generated hypergraphs, confirming their effectiveness in zero-shot tabular learning.

\begin{figure}[t]
    \centering
    \includegraphics[width=0.480\textwidth]{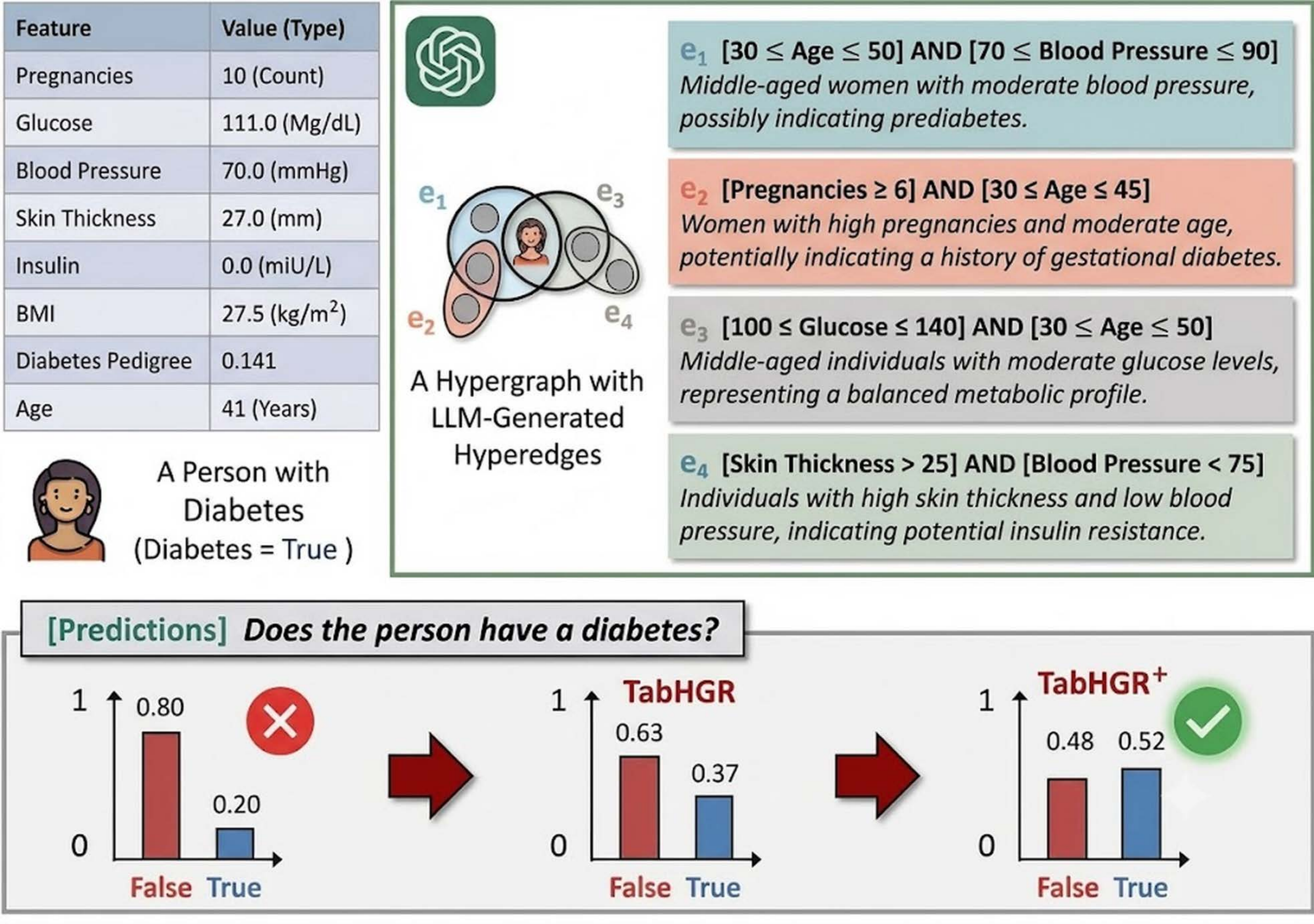} 
    \caption{
    \underline{\smash{Case study of LLM-generated hypergraphs.}}
    Given a diabetic patient, an LLM builds a hypergraph over semantically related rows. Although the initial weak label is incorrect, \naive partially corrects it via propagation, while \method emphasizes informative hyperedges and yields the correct prediction.
    \label{fig:case_study}
    }
\end{figure}

\smallsection{Case Study.}
We present a case study illustrating how LLM-generated hypergraphs support zero-shot reasoning. As shown in Figure~\ref{fig:case_study}, for a diabetic patient, the LLM constructs a hypergraph where the patient belongs to multiple semantic groups. Although the initial weak labels are incorrect, \naive partially refines the prediction via hypergraph propagation, while \method further improves it by weighting hyperedges according to their weak-label-specific effectiveness.


\subsection{Q4. Inductive Applicability}
\label{sec:exp:inductive}

While \naive and \method primarily operate transductively, their generated weak-labeling and semantic-group rules can also be applied to unseen rows. We construct a reference hypergraph from available unlabeled rows and independently insert each target row based on its rule assignments and semantic-group memberships. Weak labels are then propagated on the augmented hypergraph without accessing other target rows (Appendix~\ref{app:inductive}). As shown in Table~\ref{tab:inductive-comparison}, both inductive variants outperform TABLET and weak labeling only (WLO) on average. 
Inductive \method achieves an average AUROC of 0.7431, close to its transductive counterpart (0.7509), demonstrating effectiveness beyond the transductive setting.



\subsection{Q5. Parameter Analysis}
\begin{figure}[t]
\centering

\includegraphics[width=0.5\linewidth]{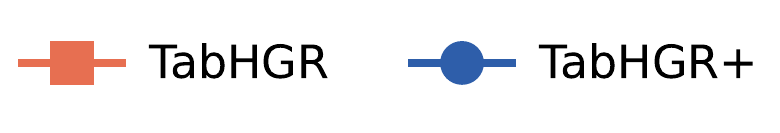}

\begin{subfigure}{\linewidth}
    \centering
    \includegraphics[width=\linewidth]{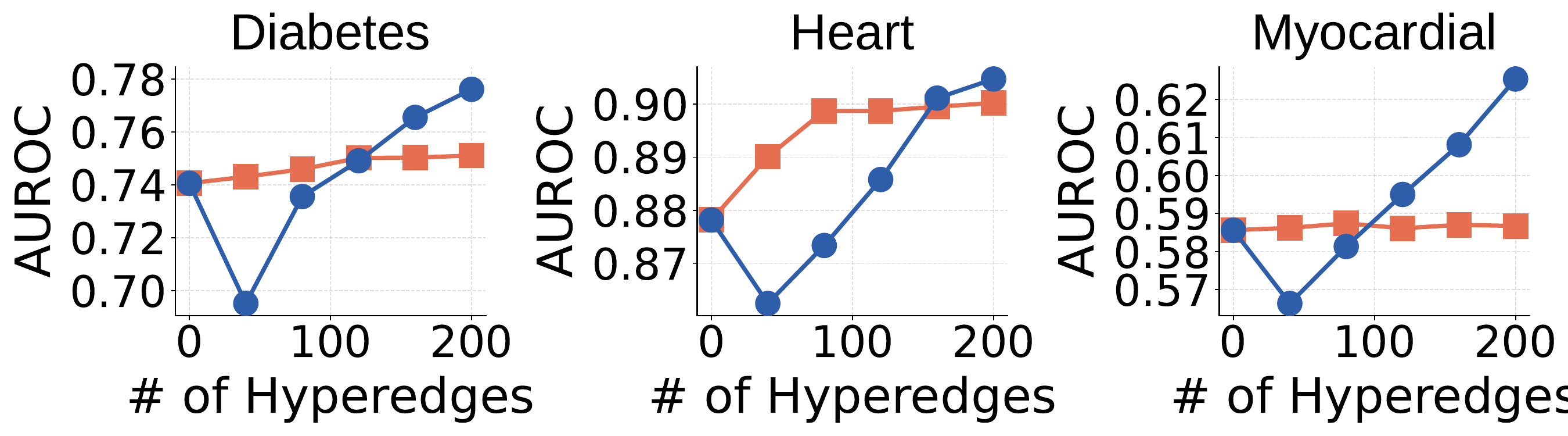}
    \caption{Number of Hyperedges}\label{fig:a}

\end{subfigure}

\vspace{7pt}

\begin{subfigure}{\linewidth}
    \centering
    \includegraphics[width=\linewidth]{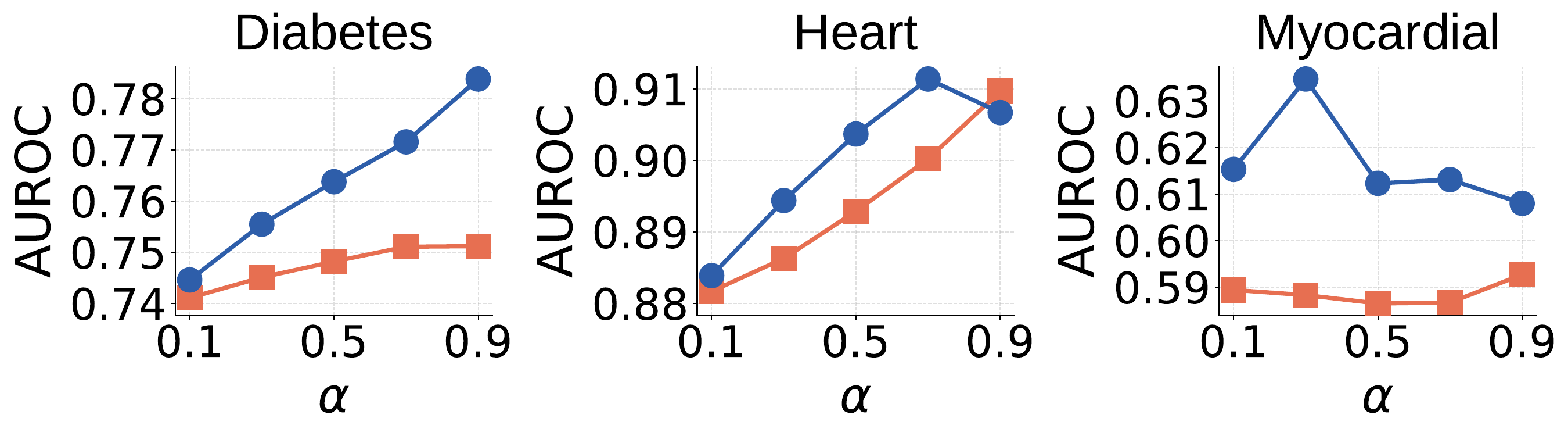}
    \caption{Propagation Strength $\alpha$}\label{fig:b}
    
\end{subfigure}

\vspace{7pt}

\begin{subfigure}{\linewidth}
    \centering
    \includegraphics[width=\linewidth]{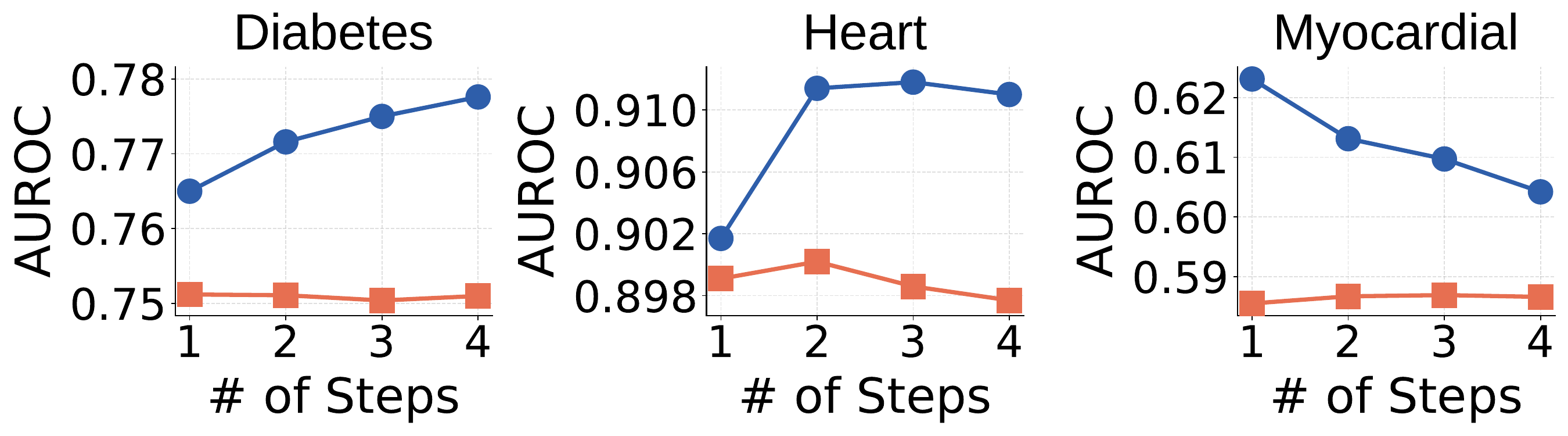}
    \caption{Number of Propagation Steps}\label{fig:c}

\end{subfigure}

\caption{
\underline{\smash{Parameter analysis.}}
The performance of \method and \naive varies with respect to (a) the number of hyperedges, (b) the propagation strength $\alpha$, and (c) the number of propagation steps. 
All parameters are fixed across datasets in the zero-shot setting.
}
\vspace{-10pt}
\label{fig:parameter}
\end{figure}

Lastly, we analyze the sensitivity of \method to key parameters: (a) the number of hyperedges, (b) the propagation strength $\alpha$, and (c) the number of propagation steps (Figure~\ref{fig:parameter}). All parameters are fixed across datasets without task-specific tuning. Increasing hyperedges generally boosts performance, yielding larger gains for \method thanks to more reliable effectiveness estimation. 
Very weak propagation (e.g., $\alpha=0.1$) degrades results. Because the optimal step count depends on hypergraph density, we leave unsupervised step adaptation to future work.


\begin{table}[t]
    \centering
\caption{
\underline{\smash{Inductive performance.}}
The inductive variants of \naive and \method process target rows independently, outperform inductive baselines on average, and remain competitive with their transductive counterparts. WLO denotes weak labeling only.
}
    \label{tab:inductive-comparison}
    \setlength{\tabcolsep}{2.0pt}
    \scalebox{0.67}{%
    \begin{tabular}{l|cccc|cc}
        \toprule
        & \multicolumn{4}{c|}{Inductive}
        & \multicolumn{2}{c}{Transductive} \\
        \cmidrule(lr){2-5}\cmidrule(lr){6-7}
        Dataset & TABLET & WLO & \naive & \method & \naive & \method \\
        \midrule
        Adult
        & 0.7723 & 0.8722 & 0.8747 & 0.8744
        & \underline{\smash{0.8817}} & \textbf{0.8820} \\

        Bank
        & 0.6910 & 0.6998 & 0.7026 & 0.7031
        & \underline{\smash{0.7092}} & \textbf{0.7202} \\

        Blood
        & 0.5139 & 0.7447 & 0.7470 & \underline{\smash{0.7545}}
        & 0.7472 & \textbf{0.7572} \\

        Car
        & 0.7785 & 0.7799 & 0.7814 & 0.7843
        & \underline{\smash{0.7900}} & \textbf{0.7966} \\

        Commun.
        & 0.6604 & 0.7700 & 0.7725 & \underline{\smash{0.7727}}
        & \underline{\smash{0.7727}} & \textbf{0.7778} \\

        Credit-G
        & 0.4681 & \textbf{0.5524} & \underline{\smash{0.5510}} & 0.5504
        & 0.5464 & 0.5285 \\

        Diabetes
        & 0.6946 & 0.7406 & 0.7413 & \underline{\smash{0.7520}}
        & 0.7511 & \textbf{0.7716} \\

        Heart
        & 0.7916 & 0.8782 & 0.8828 & 0.8943
        & \underline{\smash{0.9002}} & \textbf{0.9114} \\

        Myocar.
        & \underline{\smash{0.6103}} & 0.5856 & 0.5897 & 0.6024
        & 0.5867 & \textbf{0.6131} \\

        \midrule
        Avg.
        & 0.6645 & 0.7359 & 0.7381 & \underline{\smash{0.7431}}
        & 0.7428 & \textbf{0.7509} \\
        \bottomrule
    \end{tabular}%
    }
\end{table}

\section{Conclusions and Future Directions}
\label{sec:conc}



We propose \method, an efficient, relation-aware zero-shot tabular learning framework.
Instead of costly row-wise prompting, \method uses LLMs to generate attribute-based rules for weak labeling and semantic grouping, keeping inference costs constant regardless of dataset size. 
To refine noisy weak labels, we perform label propagation over hypergraphs formed by these LLM-inferred groups.
We also introduce label-adaptive hyperedge weighting to account for varying hyperedge contributions across labels. Empirically, \method outperforms recent zero-shot and strong few-shot baselines while substantially reducing inference costs.

A promising future direction is relational databases (RDBs)~\cite{robinson2024relbench,kim2026refuge,choi2026rdb2g}, where LLM-generated semantic hyperedges can be combined with explicit cross-table relations to support zero-shot prediction over multi-table data. 
In addition, future work could explicitly account for higher-order properties such as hyperedge overlap, degree and size distributions, and local motifs when generating and refining hypergraphs~\cite{lee2025survey}.


\section*{Limitations}
While \method achieves strong zero-shot performance with lower LLM inference cost, it has several limitations and potential risks. Its effectiveness depends on the quality of LLM-generated rules and semantic groups, which may vary across domains and prompts; biased or spurious rules could propagate to downstream predictions, especially in sensitive domains. Although LLM calls are constant with dataset size, complex datasets may require multiple prompts to generate diverse hyperedges. Fixed zero-shot hyperparameters avoid task-specific tuning but may be suboptimal under different hypergraph densities, label distributions, or distribution shifts. Finally, the framework currently focuses on classification with meaningful attribute descriptions; applying it to regression, noisy metadata, or high-stakes settings requires careful validation and human oversight.

\newpage 

\section*{Ethical Considerations}
\method relies on LLM-generated rules and semantic groups, which may contain biases or spurious correlations. These errors can be amplified through label propagation, especially in sensitive domains such as healthcare, finance, or crime prediction. Although we use public benchmark datasets, real-world deployment requires privacy protection, fairness auditing, and compliance with data-use policies. \method should thus be used as a decision-support tool with human oversight. 

\section*{Acknowledgements.}
This work was partly supported by the National Research Foundation of Korea (NRF) grant funded by the Korea government (MSIT) (No. RS-2024-00406985, 30\%).
This work was partly supported by Institute of Information \& Communications Technology Planning \& Evaluation (IITP) grant funded by the Korea government (MSIT) (No. RS-2024-00438638, EntireDB2AI: Foundations and Software for Comprehensive Deep Representation Learning and Prediction on Entire Relational Databases, 50\%)
(No. RS-2019-II190075, Artificial Intelligence Graduate School Program (KAIST), 10\%). This research was partly supported by the ``Advanced GPU Utilization Support Program'' funded by the Government of the Republic of Korea (Ministry of Science and ICT) (05-26-04-0100, 10\%). 
    
\bibliography{ref}

\newpage
\appendix
\section{Details on Experimental Settings}\label{app:exp_settings}
We provide details on the experimental settings. 

\smallsection{Datasets.}
We use nine public datasets:
(1) Adult~\cite{asuncion2007uci}, aimed at predicting whether a person’s annual income exceeds \$50K;
(2) Bank~\cite{moro2014data}, aimed at predicting a client’s likelihood of subscribing to a term deposit;
(3) Blood~\cite{yeh2009knowledge}, aimed at predicting whether blood donors will return for future donations;
(4) Car~\cite{kadra2021well}, aimed at classifying automobile quality levels;
(5) Credit-G~\cite{kadra2021well}, aimed at classifying an individual’s credit risk as good or bad;
(6) Diabetes~\footnote{\tiny\url{kaggle.com/datasets/uciml/pima-indians-diabetes-database}}, aimed at predicting the presence of diabetes based on clinical indicators;
(7) Heart\footnote{\tiny\url{kaggle.com/datasets/fedesoriano/heart-failure-prediction}}, aimed at predicting the likelihood of coronary artery disease;
(8) Communities~\cite{redmond2009communities}, aimed at predicting the crime rate of a given community; and
(9) Myocardial~\cite{golovenkin2020myocardial}, aimed at predicting whether a patient develops chronic heart failure following myocardial infarction.

\smallsection{Preprocessing Datasets with Many Attributes.}
For Communities and Myocardial, whose number of attributes exceeds 100, we perform an attribute selection step to reduce prompt complexity.
Specifically, we prompt the LLM to identify the 20 most relevant attributes based on the task description and column semantics, and use only these attributes for subsequent rule generation.
The prompt template used for attribute selection is shown in Figure~\ref{box:attribute_selection}.

\smallsection{Documentation of Artifacts.}
We provide the code and datasets used in our experiments in the anonymous repository, including the implementation of \method, LLM prompt scripts, hypergraph construction and propagation code, and scripts for reproducing the results. Our experiments use nine public tabular classification datasets: Adult, Bank, Blood, Car, Credit-G, Diabetes, Heart, Communities, and Myocardial, covering domains such as income prediction, marketing, healthcare, credit assessment, automobile evaluation, and community-level crime modeling. The artifacts are intended for academic research and reproducibility; since some datasets involve sensitive domains, users should consider potential biases and demographic coverage before applying them beyond benchmarking.

\smallsection{Data Consent.}
We use only publicly available benchmark datasets and collect no new human-subject data. We therefore rely on the original providers for consent, data-collection procedures, and usage terms, and use the datasets solely for academic benchmarking.

\smallsection{Offensive Content.}
We use only publicly available benchmark datasets and do not collect new data. We did not add personally identifying information or offensive content during artifact creation. To the best of our knowledge, the datasets do not contain direct identifiers such as names or contact information; however, some may include sensitive demographic or health-related attributes. We therefore use them only for academic benchmarking and do not attempt to identify individuals or make real-world decisions based on the data.

\smallsection{Parameters for Packages.}
\label{app:package_parameters}
We report the main implementation and parameter settings used in our experiments. For \method, all hyperparameters are fixed across datasets in the zero-shot setting, including the number of weak-labeling rules, number of semantic groups, propagation coefficient, propagation steps, and confidence ratio.
For baseline methods and external packages, we follow the official implementations and recommended settings whenever available, and use consistent preprocessing and evaluation protocols across datasets.

\section{Prompt Templates}\label{app:prompt}
We provide the prompt templates used in \method, specifically (1) weak label generation (Figure~\ref{box:rule_prompt}) and (2) group generation (Figure~\ref{box:group_prompt}).


\section{Complexity Analysis}
\label{appendix:complexity}

We analyze the time complexity of \method. 

\smallsection{Weak Label \& Hypergraph Generation.}
\method employs LLMs to generate $|\mathcal{R}|$ rules for defining weak labels and $|\mathcal{E}|$ rules for constructing hyperedges. 
In practice, we generate rules and hyperedges in batches of size $B$, requiring $\lceil |\mathcal{R}| /B\rceil$ and $\lceil |\mathcal{E}| /B\rceil$ LLM prompts, respectively. 
Thus, the total number of LLM API calls is $\lceil |\mathcal{R}| /B\rceil$ + $\lceil |\mathcal{E}| /B\rceil$, which is typically substantially smaller than row-wise prompting that requires $|\mathcal{D}|$ calls.

\smallsection{Hypergraph Propagation.}
The time complexity of a single hypergraph propagation step is $O(\sum_{e\in\mathcal{E}}|e|)$ using sparse matrix multiplications. 
Propagating for $T$ steps takes $O(T \sum_{e\in\mathcal{E}}|e|)$ time.
Since \naive propagates $|\mathcal{R}|$ weak labels, its time complexity is $O(|\mathcal{R}|\,T\sum_{e\in\mathcal{E}}|e|)$.
A hypergraph propagation consists only of forward passes and does not require gradient-based optimization.

\smallsection{Hyperedge Effectiveness Estimation.}
\method estimates hyperedge effectiveness by evaluating the impact of removing each hyperedge. 
Specifically, for each weak label, propagation is performed once on the full hypergraph and once for each hyperedge removal.
This results in $|\mathcal{E}|+1$ propagations per weak label.
Therefore, the total time complexity of \method is $O(|\mathcal{R}|\,T\,|\mathcal{E}|\sum_{e\in\mathcal{E}}|e|)$.
In practice, this step incurs negligible overhead to LLM API calls, as empirically shown in Section~\ref{sec:exp:efficiency}.

\begin{table}[t]
  \centering
  \caption{Effectiveness of hyperedge generation in AUROC. Weak-HG constructs hyperedges from weak-labeling-rule memberships, whereas \naive constructs them from LLM-generated semantic groups. Semantic hyperedges lead to better average performance.}
  \label{tab:hyperedge-source-ablation}
  \setlength{\tabcolsep}{24.8pt}
  \renewcommand{\arraystretch}{1.04}
  \scalebox{0.775}{%
  \begin{tabular}{@{}lcc@{}}
    \toprule
    Dataset & TabHGR & Weak-HG \\
    \midrule
    Adult       & \textbf{0.8817} {\small $\pm$ \textbf{0.0045}} & 0.8774 {\small $\pm$ 0.0066} \\
    Bank        & \textbf{0.7092} {\small $\pm$ \textbf{0.0157}} & 0.7013 {\small $\pm$ 0.0160} \\
    Blood       & \textbf{0.7472} {\small $\pm$ \textbf{0.0286}} & 0.7346 {\small $\pm$ 0.0367} \\
    Car         & \textbf{0.7900} {\small $\pm$ \textbf{0.0295}} & 0.7857 {\small $\pm$ 0.0312} \\
    Commun. & 0.7727 {\small $\pm$ 0.0195} & \textbf{0.7759} {\small $\pm$ \textbf{0.0210}} \\
    Credit-G    & 0.5464 {\small $\pm$ 0.0218} & \textbf{0.5513} {\small $\pm$ \textbf{0.0203}} \\
    Diabetes    & \textbf{0.7511} {\small $\pm$ \textbf{0.0131}} & 0.7450 {\small $\pm$ 0.0212} \\
    Heart       & \textbf{0.9002} {\small $\pm$ \textbf{0.0015}} & 0.8843 {\small $\pm$ 0.0025} \\
    Myocar.  & \textbf{0.5867} {\small $\pm$ \textbf{0.0093}} & 0.5812 {\small $\pm$ 0.0149} \\
    \midrule
    \textbf{Average} & \textbf{0.7428} & 0.7374 \\
    \bottomrule
  \end{tabular}%
  }
\end{table}

\section{Inductive Extension}
\label{app:inductive}

The main formulation of \naive and \method is transductive, jointly propagating weak labels over all unlabeled target rows. We additionally consider an inductive setting in which each target row is predicted without accessing the other target rows. Let
$\mathcal{D}_{\mathrm{ref}}
=\{\mathbf{s}^{\mathrm{ref}}_i\}_{i=1}^{N_{\mathrm{ref}}}$
and
$\mathcal{D}_{\mathrm{tar}}
=\{\mathbf{s}^{\mathrm{tar}}_i\}_{i=1}^{N_{\mathrm{tar}}}$
denote the unlabeled reference and target sets, respectively, where each row is represented by its attribute values.

\smallsection{Reference Hypergraph.}
We generate the weak-labeling rules $\mathcal{R}$ and semantic-group rules $\mathcal{R}_{\mathrm{group}}$ using only the task and attribute descriptions and examples sampled from $\mathcal{D}_{\mathrm{ref}}$. Applying the semantic-group rules to the reference rows produces a hypergraph
$\mathcal{G}_{\mathrm{ref}}=(\mathcal{V}_{\mathrm{ref}},\mathcal{E})$
with incidence matrix
$H_{\mathrm{ref}}\in
\{0,1\}^{N_{\mathrm{ref}}\times|\mathcal{E}|}$.
For each weak-labeling rule $r_j^{(c)}$, we also obtain its assignment vector over the reference rows,
$\mathbf{x}_{j,\mathrm{ref}}^{(c)}
\in\{0,1\}^{N_{\mathrm{ref}}}$.

\begin{table}[t]
  \centering
  \caption{Structural comparison of weak-labeling and semantic-group rules. Cov.\ denotes the average percentage of rows covered by each rule, while Jaccard denotes the average pairwise Jaccard similarity between weak-labeling and semantic-group rules. }
  \label{tab:rule-structure-comparison}
  \setlength{\tabcolsep}{10.4pt}
  \renewcommand{\arraystretch}{1.04}
  \scalebox{0.775}{%
  \begin{tabular}{@{}l|rr|c@{}}
    \toprule
    Dataset & Weak Cov. & Group Cov. & Jaccard \\
    \midrule
    Adult       & 31.3\% & 9.3\%  & 0.0581 {\small $\pm$ 0.0046} \\
    Bank        & 37.6\% & 8.8\%  & 0.0631 {\small $\pm$ 0.0010} \\
    Blood       & 29.8\% & 11.3\% & 0.0935 {\small $\pm$ 0.0030} \\
    Car         & 11.2\% & 11.1\% & 0.0610 {\small $\pm$ 0.0034} \\
    Commun.     & 32.0\% & 22.5\% & 0.1405 {\small $\pm$ 0.0026} \\
    Credit-G    & 44.7\% & 14.0\% & 0.0925 {\small $\pm$ 0.0022} \\
    Diabetes    & 32.9\% & 10.5\% & 0.0828 {\small $\pm$ 0.0018} \\
    Heart       & 35.0\% & 9.3\%  & 0.0741 {\small $\pm$ 0.0018} \\
    Myocar.     & 41.3\% & 12.4\% & 0.0797 {\small $\pm$ 0.0028} \\
    \midrule
    Average     & 32.9\% & 12.1\% & 0.0828 \\
    \bottomrule
  \end{tabular}%
  }
\end{table}

\smallsection{Target-Row Inference.}
For each target row
$\mathbf{s}^{\mathrm{tar}}_i\in\mathcal{D}_{\mathrm{tar}}$,
we apply the generated rules to determine its initial weak labels and semantic-group memberships. Let
$x_{i,j}^{(c),\mathrm{tar}}\in\{0,1\}$
denote its assignment under weak-labeling rule $r_j^{(c)}$, and let
$\mathbf{h}_i^{\mathrm{tar}}\in\{0,1\}^{|\mathcal{E}|}$
denote its hyperedge-membership vector. We insert the target row into the reference hypergraph by constructing
\begin{equation*}
H_i=
\begin{bmatrix}
H_{\mathrm{ref}}\\
(\mathbf{h}_i^{\mathrm{tar}})^\top
\end{bmatrix},
\qquad
\mathbf{x}_{i,j,\mathrm{aug}}^{(c,0)}
=
\begin{bmatrix}
\mathbf{x}_{j,\mathrm{ref}}^{(c)}\\
x_{i,j}^{(c),\mathrm{tar}}
\end{bmatrix}.
\end{equation*}
We perform the propagation described in
Section~\ref{sec:method:naive} on this augmented hypergraph and retain only the propagated score corresponding to
$\mathbf{s}^{\mathrm{tar}}_i$. Its class-level scores are then obtained by aggregating the propagated weak labels as in Eq.~\eqref{eq:z}.

Each target row is inserted and processed independently and is removed before the next target row is considered. Therefore, its prediction can use the structure and weak labels of the reference rows but cannot depend on any other target row. The reference hypergraph and generated rules are reused across all target rows.

\smallsection{Inductive Variants.}
Inductive \naive performs unweighted propagation separately on each augmented hypergraph $H_i$. For inductive \method, we first estimate the effectiveness of each hyperedge using only $\mathcal{G}_{\mathrm{ref}}$. Specifically, confident reference rows provide proxy labels, and the effectiveness of a hyperedge is measured by the change in propagation quality when it is removed, following Section~\ref{sec:method:final}. The resulting label-specific weights are fixed and reused when propagating on each augmented hypergraph $H_i$. Thus, a target row affects only its own hyperedge memberships and propagation result, without influencing weight estimation or the predictions of other target rows.

\smallsection{Evaluation Protocol.}
We evaluate predictions only on $\mathcal{D}_{\mathrm{tar}}$, without using its ground-truth labels or jointly accessing its rows during inference. We repeat the experiment over five independent weak-rule and hypergraph generations and report the mean performance.

\begin{figure}[t]
    \centering
    \begin{subfigure}{0.485\linewidth}
        \centering
        \includegraphics[width=\linewidth]{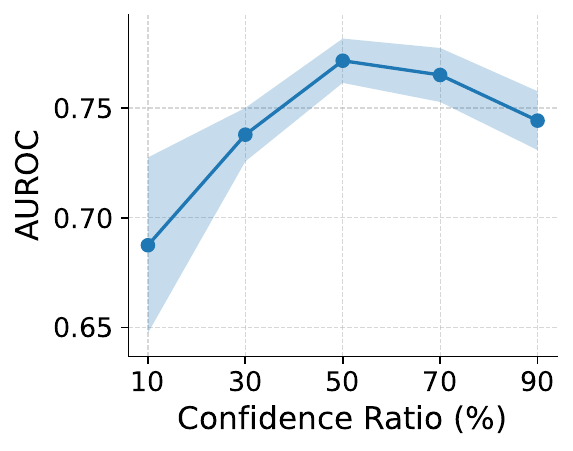}
        \caption{Diabetes}
        \label{fig:q:a}
    \end{subfigure}
    \begin{subfigure}{0.485\linewidth}
        \centering
        \includegraphics[width=\linewidth]{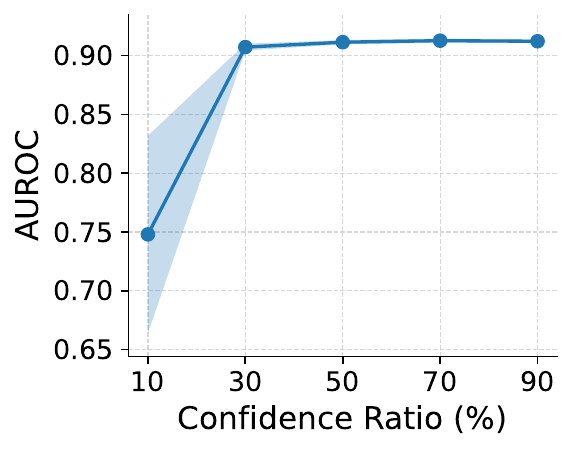}
        \caption{Heart}
        \label{fig:q:b}
    \end{subfigure}
    \caption{\method with varying confidence ratio $q$.
    Performance generally improves as $q$ increases up to $0.5$ and remains stable thereafter.}
    \label{fig:q}
    \vspace{-5pt}
\end{figure}

\section{Additional Experiments}
\label{app:additional_exp}

We provide additional experiments on the effectiveness and structural properties of LLM-generated hyperedges and the effect of the pseudo-label ratio.

\smallsection{LLM-Generated Hypergraphs.}
Table~\ref{tab:hyperedge-source-ablation} compares TabHGR with Weak-HG, a variant that replaces LLM-generated semantic-group hyperedges with hyperedges formed by samples selected by the same weak-labeling rule. TabHGR achieves a higher average AUROC (0.7428 vs.\ 0.7374) and outperforms Weak-HG on seven of nine datasets, demonstrating the effectiveness of the LLM-generated hypergraph. 
Table~\ref{tab:rule-structure-comparison} provides further structural insights. Weak-labeling rules generally have higher coverage, whereas semantic-group rules capture more fine-grained row relationships. 
Their low average pairwise Jaccard similarity further indicates that they capture largely distinct structures.

\smallsection{Effect of the Confidence Ratio.}
Figure~\ref{fig:q} shows the effect of the confidence ratio $q$. Performance improves as $q$ increases from 10 to 50 on both datasets. It slightly decreases beyond 50 on Diabetes, while remaining stable on Heart.

\begin{figure}[t]
    \centering
    \includegraphics[width=0.95\linewidth]{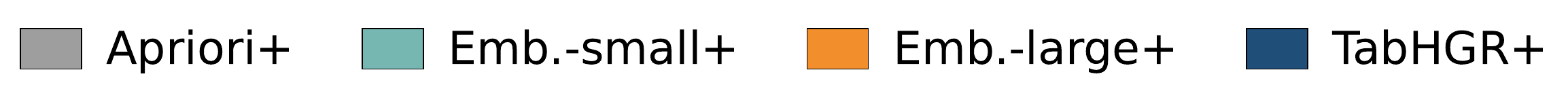}
    \begin{subfigure}{0.485\linewidth}
        \centering
        \includegraphics[width=\linewidth]{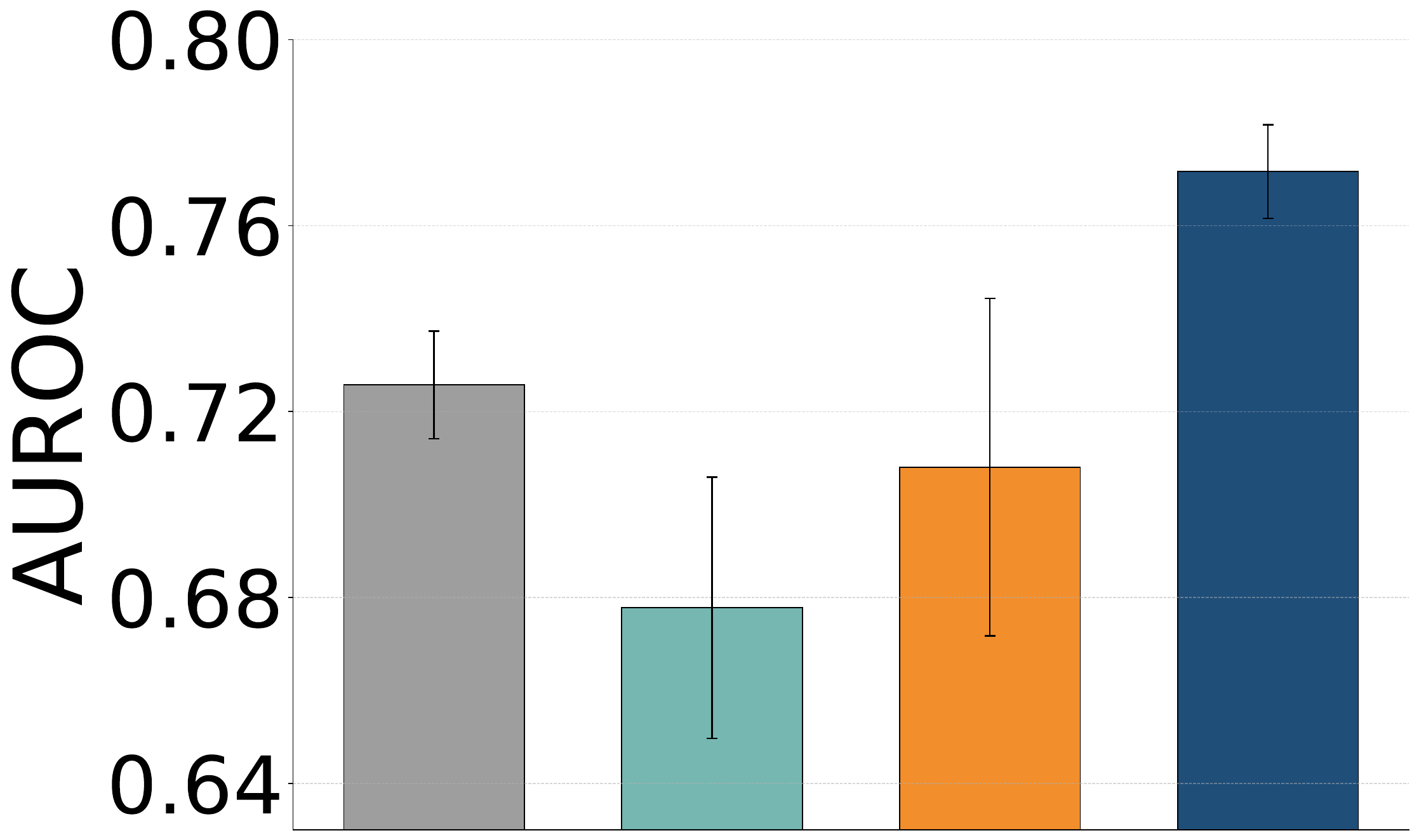}
        \caption{Diabetes}
    \end{subfigure}
    \begin{subfigure}{0.485\linewidth}
        \centering
        \includegraphics[width=\linewidth]{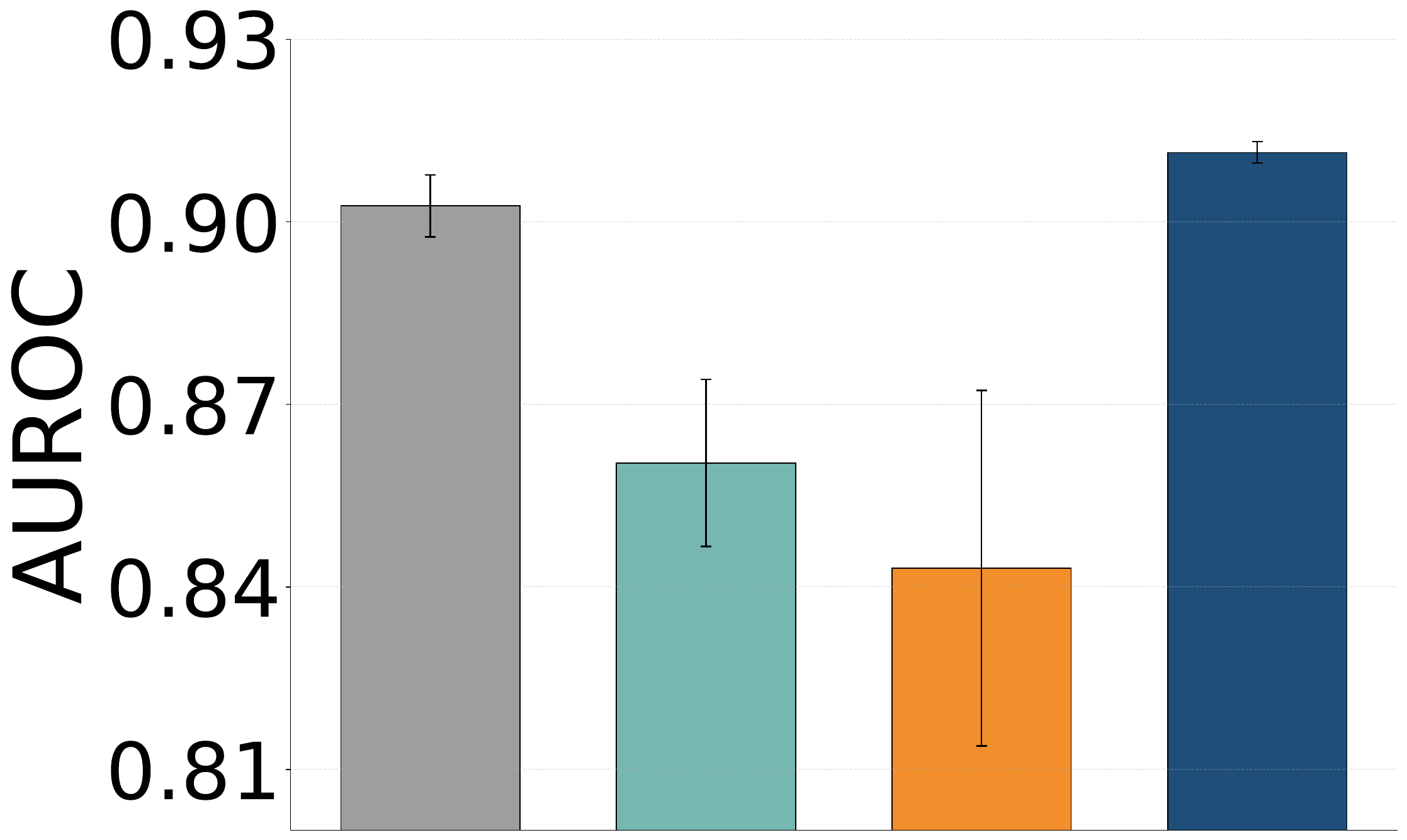}
        \caption{Heart}
    \end{subfigure}
    \caption{
    Evaluation with alternative hypergraph construction methods. 
    Apriori+ uses frequent itemsets as hyperedges, while Emb.-small+ and Emb.-large+ cluster row embeddings from \texttt{text-embedding-3-small} and \texttt{text-embedding-3-large}, respectively.
    \method achieves the best AUROC on both datasets.
    }
    \label{fig:hypergraph_baselines}
\end{figure}

\begin{figure}[t]
    \centering
    \includegraphics[width=0.7\linewidth]{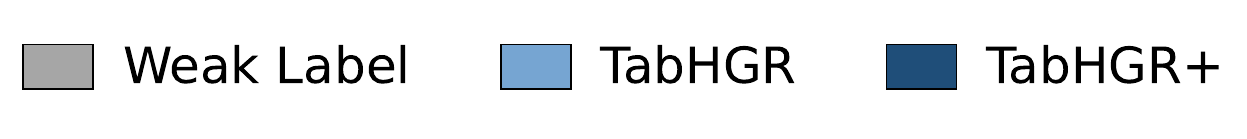}
    \begin{subfigure}{0.485\linewidth}
        \centering
        \includegraphics[width=\linewidth]{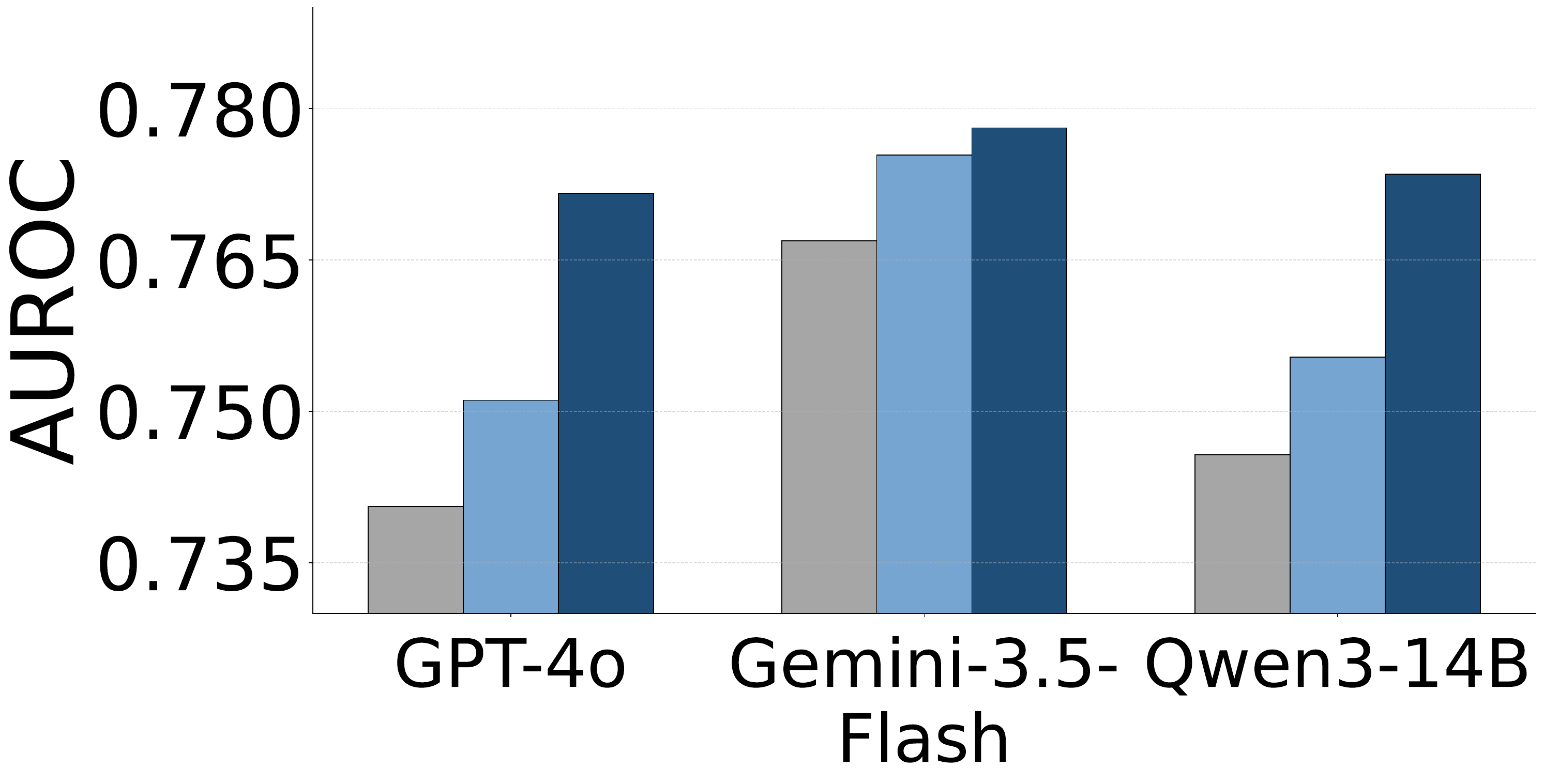}
        \caption{Diabetes}
        \label{fig:q:a}
    \end{subfigure}
    \begin{subfigure}{0.485\linewidth}
        \centering
        \includegraphics[width=\linewidth]{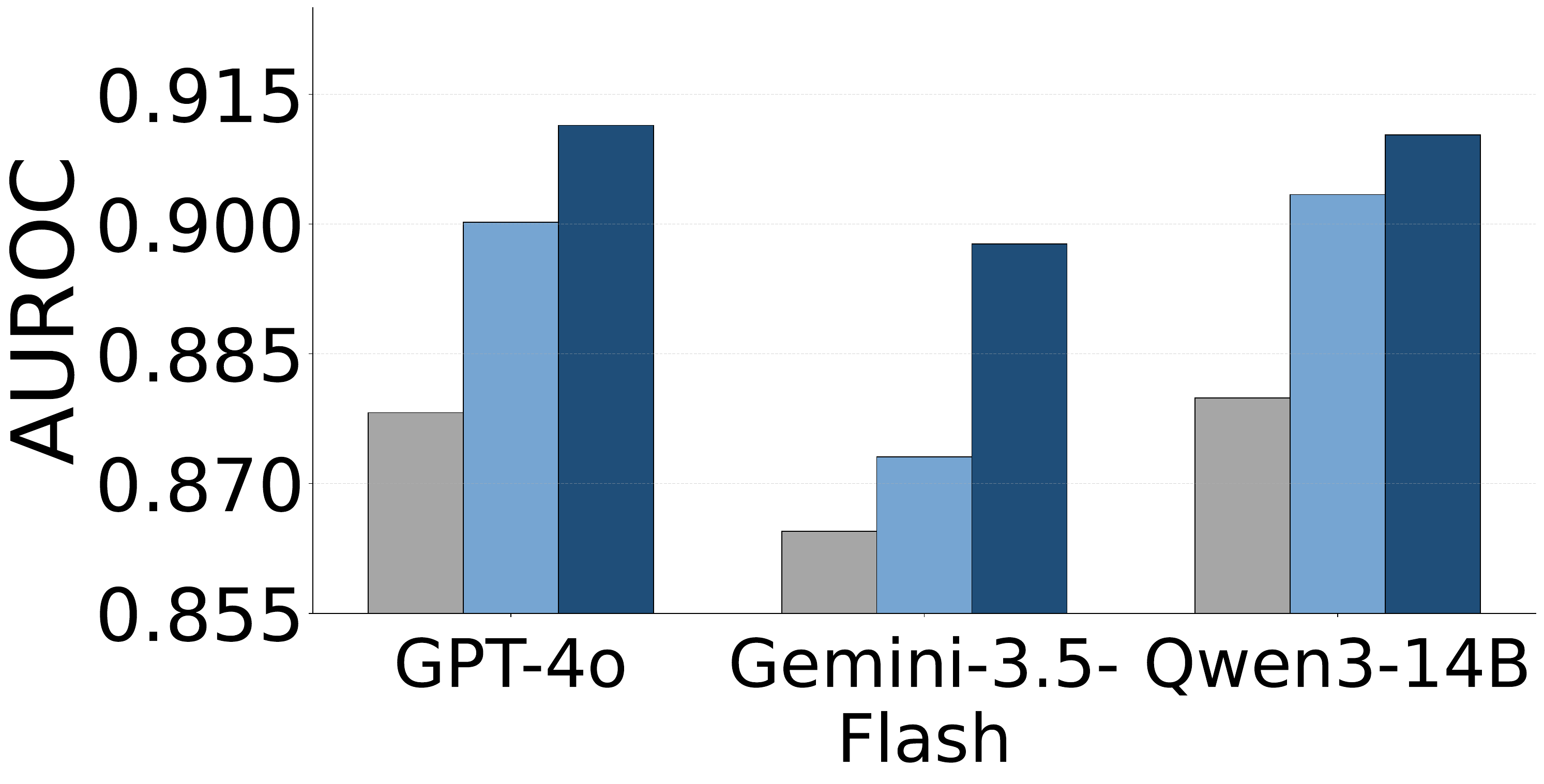}
        \caption{Heart}
        \label{fig:q:b}
    \end{subfigure}
    \caption{Performance across different LLM backbones. \naive and \method consistently improve upon weak labeling across all backbones, with \method achieving the best AUROC in every setting.}
    \label{fig:llm_backbone}
    \vspace{-5pt}
\end{figure}

\smallsection{Alternative Hypergraph Constructions.}
In addition to the results in Table~\ref{tab:llm:random_hypergraph}, we compare our LLM-generated semantic hypergraph with frequent-itemset- and embedding-based constructions. 
Apriori+~\cite{huang2000fast} discretizes continuous attributes and uses frequent itemsets as hyperedges, while Emb.-small+ and Emb.-large+ cluster row embeddings from \texttt{text-embedding-3-small} and \texttt{text-embedding-3-large}, respectively. 
As shown in Figure~\ref{fig:hypergraph_baselines}, \method consistently outperforms these alternatives on Diabetes and Heart.

\smallsection{Robustness across LLM Backbones.}
We evaluate our methods using GPT-4o~\cite{hurst2024gpt}, Gemini-3.5-Flash~\cite{googledeepmind2026gemini35}, and Qwen3-14B~\cite{yang2025qwen3} on Diabetes and Heart. 
As shown in Figure~\ref{fig:llm_backbone}, \naive consistently improves upon weak labeling, while \method yields further gains across all backbones. 
These results indicate that the benefits of hypergraph propagation and label-adaptive weighting are not specific to a particular LLM.

\begin{figure*}[t]
\centering
\begin{tcolorbox}[
    colframe=blue!50!black,
    colback=blue!3,
    title=\textbf{Feature Selection},
    fonttitle=\bfseries,
    width=\textwidth,
    boxrule=0.9pt,
    arc=2mm,
    left=4pt,
    right=4pt,
    top=4pt,
    bottom=4pt,
    sharp corners=south
]
You are a domain expert in \textbf{tabular classification}. Your objective is to \textbf{select the most useful features (columns) from the given table}.

\medskip
\textbf{Task:}  
[TASK]

\textbf{Features:}  
[FEATURES]

\textbf{Examples:}  
[EXAMPLES]

\medskip
\textbf{Instructions:}  
Select \texttt{[NUM\_FEATURE]} features that are most relevant for the classification task. Consider discriminative power, coverage, and generalizability. Think step by step.

\medskip
\textbf{Response Format (feature indices only, no explanation):}
A list of 20 feature indices separated by commas.

\end{tcolorbox}
\captionof{figure}{LLM prompt template for feature selection.}
\label{box:attribute_selection}
\end{figure*}

\begin{figure*}[t]
\centering
\begin{tcolorbox}[
    colframe=blue!50!black,
    colback=blue!3,
    title=\textbf{Class-Aware Rule Generation},
    fonttitle=\bfseries,
    width=\textwidth,
    boxrule=0.9pt,
    arc=2mm,
    left=4pt,
    right=4pt,
    top=4pt,
    bottom=4pt,
    sharp corners=south
]
You are an expert in \textbf{tabular data classification}. Your objective is to derive \textbf{interpretable and generalizable rules} that accurately characterize entities belonging to each class label.

\medskip
\textbf{Task:}  
\texttt{[TASK]}

\textbf{Features:}  
\texttt{[FEATURES]}

\textbf{Examples:}  
\texttt{[EXAMPLES]}

\medskip
\textbf{Instructions:}  
Generate \texttt{[NUM\_CONDITION]} interpretable rule conditions for each label listed in \texttt{[LABEL\_LIST]}. Each condition should be general enough to apply to unseen rows, not overly specific or narrowly tailored. Refer to the provided examples to identify useful patterns, but do not simply memorize or repeat them. Each condition should be general enough to apply to unseen rows, not overly specific to particular values in the examples. Think step-by-step. 

\medskip
For each label in \texttt{[LABEL\_LIST]} (order strictly enforced), format for its response:
\vspace{-5pt}
\begin{verbatim}
Label: [Label name or value]
[Condition 1]; [A concise yet comprehensive natural-language description]
[Condition 2]; [A concise yet comprehensive natural-language description]
...
\end{verbatim}

\medskip
\textbf{Format for [Condition]:}
\vspace{-5pt}
\begin{verbatim}
For the categorical variable only,
- [Feature_name] is in [list of Categorical_values]
For the numerical variable only,
- [Feature_name] (> or >= or < or <=) [Numerical_value]
- [Feature_name] is within range of [Numerical_range_start, Numerical_range_end]    
\end{verbatim}
\end{tcolorbox}
\captionof{figure}{LLM prompt template for interpretable rule generation.}
\label{box:rule_prompt}
\end{figure*}

\begin{figure*}[t]
\centering
\begin{tcolorbox}[
    colframe=blue!50!black,
    colback=blue!3,
    title=\textbf{Group Generation},
    fonttitle=\bfseries,
    width=\textwidth,
    boxrule=0.9pt,
    arc=2mm,
    left=4pt,
    right=4pt,
    top=4pt,
    bottom=4pt,
    sharp corners=south
]
You are an expert in \textbf{tabular data classification}. 
Your objective is to define \textbf{meaningful and interpretable groups of entities (rows)}.

\medskip

\textbf{Features:}  
\texttt{[FEATURES]}

\textbf{Examples:}  
\texttt{[EXAMPLES]}

\medskip
\textbf{Instructions:}  
Generate \texttt{[NUM\_GROUP]} interpretable and meaningful groups. Ensure groups are non-trivial (avoid overly generic or overly specific conditions). Refer to the provided examples to identify useful patterns, but do not simply memorize or repeat them. Each group should be general enough to apply to unseen rows, not overly specific to particular values in the examples. Think step-by-step. 

\medskip
Each group must:
\vspace{-5pt}
\begin{verbatim}
- Consist of 1 to 4 simple, interpretable conditions
- Be distinct from other groups (avoid overlap in definition)
- Optionally overlap in coverage (i.e., entities can belong to multiple groups)
\end{verbatim}

\medskip
Include a concise explanation that describes:
\vspace{-5pt}
\begin{verbatim}
- What kind of entities the group represents
- Their distinctive characteristics or behaviors
- How these characteristics relate to the task and target labels 
  (e.g., why this group tends to belong to or differ from the target class).  
- Why this group is meaningful or relevant based on the features
- Avoid any repetition, commentary, or meta-explanation
\end{verbatim}

\medskip
Format for response:
\vspace{-5pt}
\begin{verbatim}
Group 1: [A concise yet comprehensive natural-language description]
[Condition 1]
[Condition 2]
...

Group 2: ...
...
Group [NUM_GROUP]: ...
...
\end{verbatim}

\medskip
\textbf{Format for [Condition]:}
\vspace{-5pt}
\begin{verbatim}
For the categorical variable only,
- [Feature_name] is in [list of Categorical_values]
For the numerical variable only,
- [Feature_name] (> or >= or < or <=) [Numerical_value]
- [Feature_name] is within range of [Numerical_range_start, Numerical_range_end]    
\end{verbatim}
\end{tcolorbox}
\captionof{figure}{LLM prompt template for group generation.}
\label{box:group_prompt}
\end{figure*}

\section{Use of AI Assistants}
\label{app:ai_assistants}

We used AI assistants only for limited writing support, such as improving grammar, clarity, and conciseness of the manuscript. All research ideas, method design, experiments, analyses, and final claims were developed and verified by the authors.

\end{document}